\documentclass[numbers]{article}
\newcommand{\equalcontrib}{\thanks{Equal contribution}}
\newcommand{\corrauthor}{\thanks{Corresponding author}}

    \usepackage[preprint]{neurips_2025}

\usepackage{wrapfig}
\usepackage{graphicx}
\usepackage{amsmath}
\usepackage{amsfonts}       
\usepackage{amsthm}
\newtheorem{theorem}{Theorem}
\usepackage{amssymb}
\usepackage{enumitem}
\usepackage[T1]{fontenc}    
\usepackage{hyperref}       
\usepackage{url}            
\usepackage{caption}
\usepackage{booktabs}       
\usepackage{amsfonts}       
\usepackage{nicefrac}       
\usepackage{microtype}      
\usepackage[table]{xcolor}
\definecolor{relategreen}{RGB}{230, 255, 230}
\usepackage{multirow}
\usepackage{placeins}        
\usepackage{subcaption}
\usepackage{amsthm}
\usepackage{xspace}
\newtheorem{lemma}{Lemma}
\newcommand{\methodName}{{\sc RelatE}\xspace}
\usepackage{tikz}
\usetikzlibrary{arrows.meta, positioning, calc}
\usepackage{tikz}
\usepackage{amsmath}
\usepackage[font=footnotesize]{caption}
\usetikzlibrary{arrows.meta, decorations.pathmorphing, angles, quotes}

\title{Is Architectural Complexity Overrated?
Competitive and Interpretable Knowledge Graph Completion with \methodName}

\author{%
  \textbf{Abhijit Chakraborty}\corrauthor\hspace{0.5em}\textsuperscript{1} \and
  \textbf{Chahana Dahal}\equalcontrib\hspace{0.5em}\textsuperscript{2} \and
  \textbf{Ashutosh Balasubramaniam}\footnotemark[2]\hspace{0.5em}\textsuperscript{3} \and
  \textbf{Tejas Anvekar}\textsuperscript{1} \and
  \textbf{Vivek Gupta}\footnotemark[1]\hspace{0.5em}\textsuperscript{1} \\[4pt]
  \textsuperscript{1}Arizona State University \qquad
  \textsuperscript{2}Westminster University \qquad
  \textsuperscript{3}IIT Guwahati\\[2pt]
  {\tt \{achakr40,vgupt140\}@asu.edu}
}

\begin{document}
\maketitle
\begin{abstract}
We revisit the efficacy of simple, real-valued embedding models for knowledge graph completion and introduce \methodName, an interpretable and modular method that efficiently integrates dual representations for entities and relations. \methodName employs a real-valued phase-modulus decomposition, leveraging sinusoidal phase alignments to encode relational patterns such as symmetry, inversion, and composition. In contrast to recent approaches based on complex-valued embeddings or deep neural architectures, \methodName preserves architectural simplicity while achieving competitive or superior performance on standard benchmarks. Empirically, \methodName outperforms prior methods across several datasets: on YAGO3-10, it achieves an MRR of 0.521 and Hit@10 of 0.680, surpassing all baselines. Additionally, \methodName offers significant efficiency gains, reducing training time by 24\%, inference latency by 31\%, and peak GPU memory usage by 22\% compared to RotatE. Perturbation studies demonstrate improved robustness, with MRR degradation reduced by up to 61\% relative to TransE and by up to 19\% compared to RotatE under structural edits such as edge removals and relation swaps. Formal analysis further establishes the model’s full expressiveness and its capacity to represent essential first-order logical inference patterns. These results position \methodName as a scalable and interpretable alternative to more complex architectures for knowledge graph completion.
\end{abstract}
\section{Introduction}
The world is rich with structured knowledge entities, relationships, and facts that can be captured succinctly through knowledge graphs (KGs). These graphs, composed of triples $(e_s, r, e_o)$ where $e_s$ and $e_o$ are head and tail entities connected by relation $r$, serve as a backbone for applications in semantic search, question answering, and commonsense reasoning. However, most real-world KGs are notoriously incomplete, motivating the task of link prediction, or knowledge graph completion (KGC).
A dominant paradigm is Knowledge Graph Embedding (KGE), where entities and relations are encoded in vector spaces and scored for plausibility. From tensor decompositions (e.g., RESCAL, ComplEx~\cite{trouillon2016complex}) to neural models (e.g., ConvE~\cite{dettmers2018convolutional}) and geometric approaches (e.g., RotatE~\cite{sun2019rotate}), increasing model expressivity has often come at the cost of complexity, opaque reasoning, and computational burden.
Motivated by recent work such as \textit{“You Can Teach an Old Dog New Tricks!”}~\cite{Ruffinelli2020You}, which shows that training strategy can outweigh architectural sophistication, we revisit a fundamental question:
\emph{How far can simple, real-valued embedding models go in modern KGC?}
We introduce \methodName, a fully real-valued, modular, and interpretable embedding model for knowledge graph completion. Unlike prior complex-valued or neural approaches, \methodName combines dual-role entity and relation embeddings through phase-modulus decomposition, sinusoidal phase alignments to capture symmetry, inversion, and composition, self-adversarial negative sampling for robust optimization, and lightweight, learnable type constraints that enhance inductive bias all without sacrificing scalability or expressiveness.
Despite its simplicity, \methodName achieves strong empirical results. It outperforms RotatE on YAGO3-10 (MRR: 0.521, Hit@10: 0.680), reduces inference latency by 31\%, and shows up to 61\% less MRR degradation under structural perturbations compared to TransE. These results demonstrate that architectural minimalism when paired with principled training can achieve both efficiency and robustness.
In this paper:
\begin{itemize}
\item We introduce \methodName, a fully real-valued, interpretable knowledge graph embedding model that uses dual-role entity and relation representations with phase-modulus decomposition.
\item In \methodName we propose a sinusoidal phase alignment mechanism to capture symmetry, inversion, and composition, and integrate lightweight type-informed constraints to improve inductive bias.
\item We demonstrate that \methodName achieves competitive or superior results across FB15k-237, WN18RR, and YAGO3-10 achieving an MRR of 0.521 and Hit@10 of 0.680 on YAGO3-10, while reducing inference latency by 31\% compared to \textsc{RotatE}.
\item We conduct comprehensive ablation and perturbation studies, showing that \methodName maintains robustness under structural edits (e.g., edge removals), with up to 61\% lower MRR degradation relative to \textsc{TransE}.
\item Through both formal analysis and empirical evaluation, we establish that \methodName effectively captures key first-order inference patterns, reinforcing that minimalist architectures can be both scalable and expressive.
\end{itemize}
Our findings highlight an underexplored design principle: that simplicity augmented with modern training techniques can rival architectural complexity in both performance and interpretability.
\begin{figure}[h]
\centering
\resizebox{\textwidth}{!}{%
\begin{tikzpicture}[>=Stealth, every node/.style={font=\small}]

\begin{scope}[xshift=0cm]
\draw[->] (0,0) -- (2.5,0) node[anchor=west] {Modulus};
\draw[->] (0,0) -- (0,2.2) node[anchor=south east] {Phase};
\draw[->, very thick, red] (0,0) -- (2,1.2) node[pos=0.6, above right] {$r$};
\draw[dashed] (2,1.2) -- (2,0);
\draw[dashed] (2,1.2) -- (0,1.2);
\draw (0.6,0) arc[start angle=0, end angle=31, radius=0.6];
\node at (0.85,0.15) {$\theta_r$};
\draw[->, thick] (2.3,2.0) to[out=210, in=50] (1.5,1.0);
\node[align=left, anchor=west, font=\scriptsize] at (2.4,2.0) {Sinusoidal\\Phase scoring};
\node[align=center, font=\bfseries] at (1,-0.6) {Phase-Modulus\\Decomposition};
\end{scope}

\begin{scope}[xshift=5.5cm]
\node[circle, draw] (a) at (0,1.2) {$e_1$};
\node[circle, draw] (b) at (2,1.2) {$e_2$};
\draw[->, thick, blue] (a) -- node[above] {\textcolor{blue}{$r$}} (b);
\draw[->, thick, dashed, blue] (b) to[bend left=40] node[below] {\textcolor{blue}{$r$}} (a);
\node at (1,-0.1) {\textbf{Symmetry}};
\end{scope}

\begin{scope}[xshift=10cm]
\node[circle, draw] (c1) at (0,1.2) {$e_1$};
\node[circle, draw] (c2) at (2,1.2) {$e_2$};
\draw[->, thick, red] (c1) -- node[above] {\textcolor{red}{$r_1$}} (c2);
\draw[->, thick, dashed, red] (c2) to[bend left=40] node[below] {\textcolor{red}{$r_2 = -r_1$}} (c1);
\node at (1,-0.1) {\textbf{Inversion}};
\end{scope}

\begin{scope}[xshift=15cm]
\node[circle, draw] (e1) at (0,1.2) {$e_1$};
\node[circle, draw] (e2) at (1.5,1.2) {$e_2$};
\node[circle, draw] (e3) at (3,1.2) {$e_3$};
\draw[->, thick, orange] (e1) -- node[above] {\textcolor{orange}{$r_1$}} (e2);
\draw[->, thick, orange!70!black] (e2) -- node[above] {\textcolor{orange!70!black}{$r_2$}} (e3);
\draw[->, thick, dashed, purple] (e1) to[bend right=35] node[below] {\textcolor{purple}{$r_3 = r_1 + r_2$}} (e3);
\node at (1.5,-0.1) {\textbf{Composition}};
\end{scope}
\end{tikzpicture}
}
\caption{ \small \methodName’s phase-modulus decomposition and its modeling of symmetry, inversion, and composition.}
\label{fig:relate-patterns}
\end{figure}
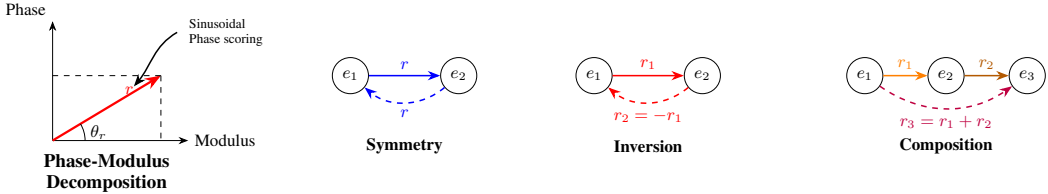
\FloatBarrier  
\vspace{-1.8em}
\section{Knowledge Base Completion: Problem, Properties, and Evaluation}
Knowledge graphs (KGs) represent structured knowledge as collections of triples $(e_h, r, e_t)$, where $e_h$ and $e_t$ are entities and $r$ is a binary relation. The task of knowledge base completion (KBC) is to infer plausible but unobserved triples from an incomplete KG. In our setting, a knowledge base (KB) is a finite set of such triples, and KBC is framed as a link prediction task over these relational facts using real-valued embeddings.
Formally, let $E$ and $R$ denote the finite sets of entities and relations, respectively. A fact is a tuple $r(e_h, e_t) \in R \times E \times E$. Given a set of training triples, the model learns to assign higher plausibility scores to true facts than to corrupted variants, which are generated by replacing either the head or tail entity with a random entity not appearing in the triple. Performance is evaluated using standard metrics such as Mean Rank (MR), Mean Reciprocal Rank (MRR), and Hits@K, following the filtered ranking protocol of \cite{NIPS2013_1cecc7a7} on standard datasets such as FB15k-237, WN18RR, and YAGO3-10. Beyond empirical accuracy, models are often also evaluated along two parallel dimensions:


\textbf{(1) Expressiveness}. A KBC model is fully expressive if it can assign arbitrary scores to any set of true and false triples.

\textbf{(2) Inference Patterns}. Logical patterns such as symmetry, inversion, and composition enable systematic generalization from observed triples. \methodName captures these as follows (see Figure~\ref{fig:relate-patterns}): \textbf{(a) Symmetry}: For relations like \textit{is\_sibling\_of}, where $r(e_1, e_2)$ implies $r(e_2, e_1)$, score should be the same i.e. invariant, \textbf{(b) Inversion}: For inverse relation pairs (e.g., \textit{born\_in} and \textit{birthplace\_of}), the model should ensure inverse relations, $r^{-1} = -r$ without duplicating the embeddings of the relation, and \textbf{(c) Composition}: For relation chains like $r_1(e_1, e_2)$ and $r_2(e_2, e_3)$, the model should be able to compose, i.e. implying $r_3(e_1, e_3)$ (e.g., \textit{grandparent}).

\textbf{Why \methodName works?}  \methodName addresses knowledge base completion (KBC) through a modular, real-valued decomposition of entities and relations into interpretable modulus and phase components. This formulation captures fine-grained relational patterns such as symmetry, inversion, and composition without relying on complex-valued arithmetic or deep neural encoders. The phase-modulus decomposition enhances expressiveness, allowing \methodName to model asymmetric, cyclic, and hierarchical structures that traditionally required complex or quaternion embeddings. Its purely real-valued parameterization not only improves scalability but also mitigates overfitting. Empirically, \methodName shows strong performance on all FB15k-237, WN18RR, and YAGO3-10.

\methodName also maintains score invariance under entity swapping by consistently aligning phase and modulus representations. Inversion is naturally handled through phase negation, ensuring that $r^{-1} = -r$ without duplicating relation embeddings. Composition is modeled via additive operations on both phase and modulus vectors, enabling accurate approximation of composite relations. Figure~\ref{fig:relate-patterns} provides a visual summary of these decomposition principles and their role in logical reasoning. By embedding rich relational structures within a lightweight and interpretable framework, \methodName challenges the assumption that architectural complexity is necessary for expressive reasoning in knowledge graph models. More details on \methodName are provided in the subsequent section \ref{sec:methodDetails}.



\section{The \methodName Model}
\label{sec:methodDetails}
We now describe \methodName, detailing its representational structure and inductive biases. \methodName embeds entities and relations in real-valued phase and modulus components, enabling interpretable reasoning over directional and scalar interactions.

\textbf{Representation.} Each entity $e \in \mathcal{E}$ and each relation $r \in \mathcal{R}$ is represented as a pair of vectors in $\mathbb{R}^{d/2}$: a phase vector $e^{(p)}, r^{(p)}$ and a modulus vector $e^{(m)}, r^{(m)}$. Additionally, each relation $r$ is associated with a bias vector $b_r \in \mathbb{R}^{d/2}$ that refines its scaling influence. The phase components encode directional or cyclical semantics (e.g., inversion, symmetry), while modulus components regulate strength and hierarchy.
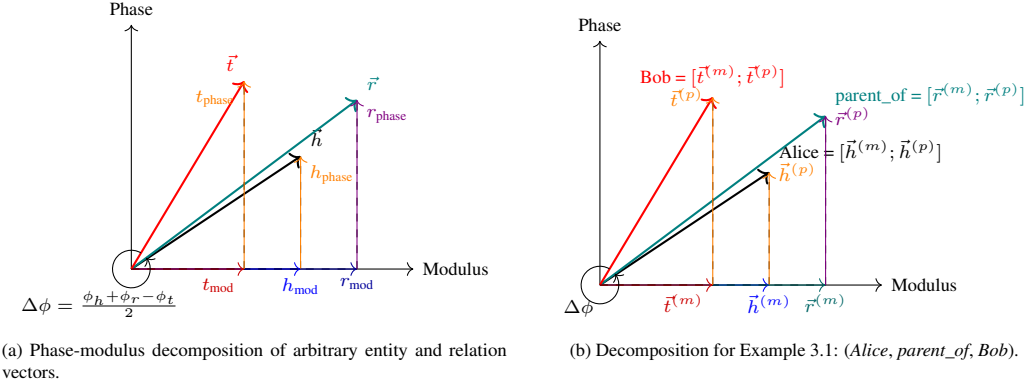
\begin{figure}[t]
\centering

\begin{subfigure}[t]{0.48\textwidth}
\centering
\begin{tikzpicture}[scale=2.5, every node/.style={font=\scriptsize}]
  \coordinate (O) at (0,0);
  \draw[->] (0,0) -- (1.5,0) node[right] {Modulus};
  \draw[->] (0,0) -- (0,1.3) node[above] {Phase};

  \coordinate (H) at (0.9,0.6);
  \draw[->, thick] (0,0) -- (H) node[above right] {$\vec{h}$};
  \draw[dashed] (0,0) -- (0.9,0);
  \draw[dashed] (0.9,0) -- (H);
  \draw[->, blue] (0,0) -- (0.9,0) node[below] {$h_{\text{mod}}$};
  \draw[->, orange] (0.9,0) -- (H) node[below right] {$h_{\text{phase}}$};

  \coordinate (R) at (1.2,0.9);
  \draw[->, thick, teal] (0,0) -- (R) node[above right] {$\vec{r}$};
  \draw[dashed] (0,0) -- (1.2,0);
  \draw[dashed] (1.2,0) -- (R);
  \draw[->, blue!60!black] (0,0) -- (1.2,0) node[below] {$r_{\text{mod}}$};
  \draw[->, violet] (1.2,0) -- (R) node[below right] {$r_{\text{phase}}$};

  \coordinate (T) at (0.6,1.0);
  \draw[->, thick, red] (0,0) -- (T) node[above left] {$\vec{t}$};
  \draw[dashed] (0,0) -- (0.6,0);
  \draw[dashed] (0.6,0) -- (T);
  \draw[->, red!80!black] (0,0) -- (0.6,0) node[below left] {$t_{\text{mod}}$};
  \draw[->, orange] (0.6,0) -- (T) node[below left] {$t_{\text{phase}}$};

  \draw pic[
    draw=black,
    ->,
    angle radius=0.25cm,
    angle eccentricity=2.5,
    "$\Delta\phi = \frac{\phi_h + \phi_r - \phi_t}{2}$"
  ] {angle=T--O--R};
\end{tikzpicture}
\caption{Phase-modulus decomposition of arbitrary entity and relation vectors.}
\label{fig:relate-decomposition}
\end{subfigure}
\hfill
\begin{subfigure}[t]{0.48\textwidth}
\centering
\begin{tikzpicture}[scale=2.5, every node/.style={font=\scriptsize}]
  \coordinate (O) at (0,0);
  \draw[->] (0,0) -- (1.5,0) node[right] {Modulus};
  \draw[->] (0,0) -- (0,1.3) node[above] {Phase};

  \coordinate (H) at (0.9,0.6);
  \draw[->, thick] (0,0) -- (H) node[above right] {Alice = $[\vec{h}^{(m)}; \vec{h}^{(p)}]$};
  \draw[dashed] (0,0) -- (0.9,0);
  \draw[dashed] (0.9,0) -- (H);
  \draw[->, blue] (0,0) -- (0.9,0) node[below] {$\vec{h}^{(m)}$};
  \draw[->, orange] (0.9,0) -- (H) node[right] {$\vec{h}^{(p)}$};

  \coordinate (R) at (1.2,0.9);
  \draw[->, thick, teal] (0,0) -- (R) node[above right] {parent\_of = $[\vec{r}^{(m)}; \vec{r}^{(p)}]$};
  \draw[dashed] (0,0) -- (1.2,0);
  \draw[dashed] (1.2,0) -- (R);
  \draw[->, teal!80!black] (0,0) -- (1.2,0) node[below] {$\vec{r}^{(m)}$};
  \draw[->, violet] (1.2,0) -- (R) node[right] {$\vec{r}^{(p)}$};

  \coordinate (T) at (0.6,1.0);
  \draw[->, thick, red] (0,0) -- (T) node[above] {Bob = $[\vec{t}^{(m)}; \vec{t}^{(p)}]$};
  \draw[dashed] (0,0) -- (0.6,0);
  \draw[dashed] (0.6,0) -- (T);
  \draw[->, red!80!black] (0,0) -- (0.6,0) node[below left] {$\vec{t}^{(m)}$};
  \draw[->, orange] (0.6,0) -- (T) node[left] {$\vec{t}^{(p)}$};

  \draw pic[
    draw=black,
    ->,
    angle radius=0.25cm,
    angle eccentricity=1.7,
    "$\Delta\phi$"
  ] {angle=T--O--R};
\end{tikzpicture}
\caption{Decomposition for Example 3.1: (\textit{Alice}, \textit{parent\_of}, \textit{Bob}).}
\label{fig:relate-example}
\end{subfigure}
\vspace{-0.2em}
\caption{2D illustrations of \textsc{RelatE}'s entity and relation decomposition into modulus and phase components. Left: abstract formulation. Right: example grounded in real-world semantics.}
\vspace{-1.0em}
\label{fig:2D_relate-joint}
\end{figure}
To compute the final representation of a triple $(h, r, t)$, the model evaluates a hybrid score based on phase difference and modulus translation. The final embedding of the head and tail relative to a relation reflects angular coherence and magnitude alignment in their respective subspaces(see figure ~\ref{fig:relate-decomposition}).

\textbf{Example 3.1.} Consider a 2D embedding configuration over a knowledge graph with the triple (\textit{Alice}, \textit{parent\_of}, \textit{Bob})  as shown in Figure \ref{fig:relate-example}. In this setup, the head entity \textit{Alice} is represented by modulus and phase vectors $\vec{h}^{(m)}$ and $\vec{h}^{(p)}$, while the tail \textit{Bob} has $\vec{t}^{(m)}$ and $\vec{t}^{(p)}$. The relation \textit{parent\_of} is encoded with $\vec{r}^{(m)}$, $\vec{r}^{(p)}$, and a learnable bias $\vec{b}_r$. To compute the score for this triple, \methodName computes the modulus translation $\vec{h}^{(m)} \circ (\vec{r}^{(m)} + \vec{b}_r)$, which moves Alice’s embedding closer to Bob’s in magnitude, where $\circ$ is the Hadamard product. Concurrently, \methodName computes the angular difference between $\vec{h}^{(p)} + \vec{r}^{(p)}$ and $\vec{t}^{(p)}$, applying a sinusoidal penalty based on their alignment. The triple receives a high score if (i) the modulus component of \textit{Alice} closely matches that of \textit{Bob} after applying the relation-specific transformation, and (ii) the angular distance between the transformed phase of \textit{Alice} and the phase of \textit{Bob} is small. This configuration reflects a true fact in the knowledge graph and demonstrates how \methodName geometrically aligns both scalar and directional relational cues. This setup can be extended to capture reflexive facts such as \textnormal{(Bob, related\_to, Bob)}, or compositional chains like \textnormal{(Alice, grandparent\_of, Charlie)} from two \textit{parent\_of} edges, leveraging additive operations across phase and modulus spaces.

\textbf{Scoring Function.}\label{scoring} In the above example, we identified plausible triples that should be scored higher to reflect the underlying relational geometry encoded by \methodName. To construct a scoring function that reflects these intuitions, we build from first principles namely, the need to assess alignment in both scalar space (modulus) and angular space (phase). Inspired by region-based frameworks like BoxE~\cite{DBLP:journals/corr/abs-2007-06267} that assign high scores when entities fall within parameterized relational regions, \methodName adopts an additive-angular (\textit{phase component}) and multiplicative-scalar (\textit{modulus component}) formulation to assess the plausibility of triples. The modulus component quantifies the scalar mismatch between the transformed head and tail entities using a learned relation-specific weight vector $w_r \in \mathbb{R}^{d/2}$. The translation includes relation-dependent scaling and type-bias interactions:
\begin{align}
\text{Modulus\ Score}(h, r, t) &= \sum_i w_r^{(i)} \cdot \left| h_i^{(m)} \cdot (r_i^{(m)} + b_i) - t_i^{(m)} \cdot (1 - b_i) \right| \label{eq:modulus}
\end{align}

This slope-weighted L1 formulation provides anisotropic sensitivity to embedding dimensions, enhancing the model’s ability to represent complex hierarchies and one-to-many mappings. For example, transitive chains such as \textit{(Alice, employed\_by, Google)} and \textit{(Google, headquartered\_in, California)} enable the inferred fact \textit{(Alice, has\_affiliation\_in, California)} to be captured by consistent modulus scaling across relation pairs. The phase component, in contrast, captures directionality. It computes the sine of the halved phase difference between $\vec{h}^{(p)} + \vec{r}^{(p)}$ and $\vec{t}^{(p)}$, encouraging minimal angular misalignment. The sinusoidal formulation encourages smooth angular continuity, while heavily penalizing disaligned or inverted directions analogous to being outside a relational region. This facilitates encoding of complex structures such as compositions (e.g., \textit{grandparent} $\approx$ \textit{parent} $\circ$ \textit{parent}) and inverse relations (e.g., \textit{born\_in} and \textit{birthplace\_of}). \methodName’s robustness under inversion and structural perturbation is empirically validated through perturbation experiments and UMAP analyses (see Appendix~\ref{sec:perturbation}, Figures~\ref{fig:umap-inversion},~\ref{fig:delta_mrr-drop})
\begin{align}
\text{Phase Score}(h, r, t) &= \left| \sin\left(\frac{h^{(p)} + r^{(p)} - t^{(p)}}{2} \right) \right|_1 \label{eq:phase}
\end{align}

This term promotes smooth transitions in angular space, preserving periodicity and discouraging disaligned or inverted phases. Hence, \methodName introduces a novel slope-weighted L1-based modulus scoring function, where each relation is associated with a learnable width vector $w_r$ that weights the element-wise differences between transformed head and tail modulus. The total score combines both components using relation-specific softplus-weighted scalars $\lambda_r^{(m)}$ and $\lambda_r^{(p)}$, and subtracts the sum from a global margin $\gamma$:
\begin{align}
f(h, r, t) &= \gamma - \left( \lambda_r^{(m)} \cdot \text{ModulusScore}(h,r,t) + \lambda_r^{(p)} \cdot \text{PhaseScore}(h,r,t) \right) \label{eq:score}
\end{align}

As defined in Equation~\eqref{eq:score}, this formulation allows each relation to dynamically prioritize phase or modulus alignment based on structural cues in the graph.

\methodName integrates a lightweight, learnable type bias inferred solely from the training and validation data (see Appendix~\ref{sec:kgc-implementation}), and injects it via a dot-product regularizer modulated by warm-start scaling enabling interpretable, type-aware reasoning without symbolic constraints.

\methodName is trained using a \emph{margin-based ranking loss}:
\begin{align*}
\mathcal{L} = \sum_{(h,r,t^+),(h,r,t^-)} \max(0, f(h, r, t^-) - f(h, r, t^+) + \gamma)
\end{align*}
where $(h,r,t^+)$ is a true triple and $(h,r,t^-)$ is a negatively sampled triple. Self-adversarial negative sampling emphasizes harder negatives. L3 regularization is applied across all parameters. To enhance generalization over sparse and asymmetric structures, \methodName augments each interaction with a latent reverse implication capturing bidirectional semantic patterns without relying on explicit inversion rules. This duality improves robustness and enables learning from low-frequency relational signals.

\methodName to the best of our knowledge is the first real-valued model to combine phase-modulus decomposition, sinusoidal and slope-weighted scoring, and lightweight type bias in a coherent, interpretable framework achieving high expressiveness without complex numbers or deep architectures.
The next section theoretically analyzes how \methodName captures key inference patterns symmetry, inversion, and composition and demonstrates its expressiveness in practice.
\section{Model Properties: Expressivity and Inductive Bias in \methodName}
Having established the architectural foundations of \methodName, we now examine its expressivity and reasoning capabilities. Its modular phase-modulus decomposition captures relational patterns and supports compositionality, constraints, and type-based inductive bias.
\subsection{Expressivity of \methodName}
We show that \methodName is fully expressive under a modular real-valued decomposition with embedding dimensionality $d = |E||R|$, enabling it to represent any arbitrary truth assignment over triples. This expressiveness arises from the ability to apply orthogonal perturbations to either the phase (direction) or modulus (magnitude) of entity embeddings, thereby providing independent control over triple plausibility. By selectively adjusting a single component, \methodName reduces the score of any false triple below zero while keeping all true triples above a fixed margin $\gamma$ achieving precise separation without relying on complex-valued embeddings or deep interaction layers.
\begin{theorem}
\methodName is fully expressive with embedding dimensionality $d = |E||R|$ under a modular, real-valued scoring function, where each entity and relation is represented via independently learned phase and modulus components. For any binary knowledge graph and any arbitrary truth assignment over triples $(h, r, t)$, there exists a configuration of real-valued embeddings such that all true triples score above a fixed margin $\gamma$, and all false triples score below zero.
\end{theorem}

\begin{lemma}\textbf{Phase–based Separation:}
\label{lem:phase}
Given any true triple $(h,r,t)$ and an arbitrary false triple $(h,r,t')$ with $t' \neq t$, there exists a perturbation 
$\delta^{(p)} \!\in\! \mathbb{R}^{d/2}$ applied \emph{only} to the phase of $t'$ such that
\[
f\!\bigl(h,r,t\bigr) \;>\; \gamma \quad\text{and}\quad 
f\!\bigl(h,r,t'\bigr) \;<\; 0,
\]
\end{lemma}
while the score of every other triple remains unchanged.

\begin{lemma} \textbf{Modulus–based Separation:}
\label{lem:mod}
Given any true triple $(h,r,t)$ and an arbitrary false triple $(h',r,t)$ with $h' \neq h$, there exists a perturbation 
$\delta^{(m)} \!\in\! \mathbb{R}^{d/2}$ applied \emph{only} to the modulus of $h'$ such that
\[
f\!\bigl(h,r,t\bigr) \;>\; \gamma \quad\text{and}\quad 
f\!\bigl(h',r,t\bigr) \;<\; 0,
\]
\end{lemma}
while the score of every other triple remains unchanged.

\begin{proof}
Enumerate all false triples and iteratively deploy Lemma~\ref{lem:phase} or Lemma~\ref{lem:mod} to each, choosing a phase or modulus perturbation that is \emph{orthogonal} to every perturbation already applied.  Because each step modifies only one embedding component and never revisits it, at most $|E||R|$ real dimensions are required.  After $|E||R|$ steps every true triple scores above the margin~$\gamma$ and every false triple scores below~$0$, establishing full expressivity for $d=|E||R|$. See appendix~\ref{sec:expressivity} for detail proof.
\end{proof}

This result positions \methodName as the first in prior art real-valued, modular, and interpretable KGE model proven to be fully expressive without relying on complex numbers, box constraints, or symbolic encodings. Its time and space complexity are $\mathcal{O}(nd)$ and $\mathcal{O}((|E| + n|R|)d)$, respectively, where $n$ denotes relation arity (with $n{=}2$ for standard triple-based graphs). These guarantees ensure tractability even on large-scale knowledge graphs. \methodName also demonstrates strong empirical scalability in both training and inference; formal complexity analysis is provided in Appendix~\ref{sec:relate-complexity}.

We show that \methodName captures a broad class of inference patterns commonly studied in knowledge graph reasoning, including symmetry, inversion, hierarchy, composition, and mutual exclusion (see formal analysis in Appendix~\ref{sec:inference_patterns}). These patterns are enabled by the model’s modular scoring function (Equation~\eqref{eq:score}), which independently parameterizes phase and modulus components for each relation allowing directionality and scalar semantics to be flexibly combined.

\begin{theorem}
\label{theorem_2}
\methodName models the following first-order relational inference patterns under its phase-modulus decomposition:

\vspace{-0.75em}
\begin{itemize}{
\setlength{\itemsep}{0.5pt}
    \item \textbf{Symmetry \& Anti-Symmetry}: Encoded via phase alignment. Symmetric relations exhibit zero phase shift, while anti-symmetric relations induce directional offsets.
    \item \textbf{Inversion}: Captured through role-reversed phase mirroring, where the phase of an inverse relation is the negative of its counterpart.
    \item \textbf{Hierarchy}: Modeled via relational modulus scaling; superclasses cover larger magnitude regions, enabling soft type generalization.
    \item \textbf{Composition}: Supported through additive phase and multiplicative modulus operations; this allows the model to infer chained relations such as $\texttt{parent} \circ \texttt{parent} \Rightarrow \texttt{grandparent}$.
    \item \textbf{Disjointness \& Mutual Exclusion}: Enforced via orthogonal phase vectors or disjoint modulus supports, preventing overlap between incompatible relations.
}\end{itemize}

\end{theorem}

\begin{proof}
\methodName encodes core inference patterns via phase-modulus decomposition: phase shifts capture symmetry, inversion, and composition, while modulus scaling models hierarchy and disjointness. Full derivations are provided in Appendix~\ref{sec:inference_patterns}.
\end{proof}
\vspace{-1.0em}
Empirical analyses in Appendix~\ref{sec:inference_patterns} and~\ref{sec:perturbation} confirm that \methodName preserves relational structure under perturbations: symmetric and inverse relations form coherent phase clusters, disjoint relations separate in modulus space, and compositional paths remain robust. UMAP visualizations (Figures~\ref{fig:umap-phase}–\ref{fig:umap-modulus}) show that \methodName preserves phase coherence for symmetric/inverse relations, maintains compositional robustness, and separates disjoint relations in modulus space.

This result highlights that \methodName despite being fully real-valued achieves expressive relational reasoning and robustness through a modular, interpretable scoring function, without resorting to complex-valued embeddings, geometric encodings, or neural parameterizations.
\section{Experimental Evaluation}
\label{sec:relate-neurips}

We evaluate \methodName on three widely used knowledge graph benchmarks FB15k-237, WN18RR, and YAGO3-10 using both uniform and self-adversarial negative sampling strategies for link prediction (KBC). Following standard KGE evaluation protocols~\cite{bordes2013translating, sun2019rotate}, all models are assessed under the filtered ranking setting with the same dataset splits. For baseline comparison, we include translational models (TransE~\cite{bordes2013translating}, RotatE~\cite{sun2019rotate}) and bilinear models (DistMult~\cite{yang2015embedding}, ComplEx~\cite{trouillon2016complex}, TuckER~\cite{balazevic2019tucker}, QuatE~\cite{zhang2019quaternion}). Baseline results are taken from published work where available; otherwise, we report our best reproduced results (\textit{italicized}) under matched experimental conditions.

To ensure fair comparison, all models use embedding dimensionality constrained to $d \leq 1000$. Higher-dimensional variants (e.g., ComplEx with $d \geq 2000$) are reported separately. \methodName is trained for up to 200K steps with early stopping based on validation MRR, and test metrics are reported from the best checkpoint. Full training details and hyperparameter configurations are provided in appendices~\ref{sec:Hyperparameter Tuning and Configuration},\ref{sec:kgc-implementation} and Table~\ref{tab:relate_hparams}.

\begin{table}[t]
  \centering
  \small
  \caption{ \small Link prediction performance of \methodName versus baselines across three benchmarks. All models are evaluated under filtered settings; overall best values are boxed.}
  \vspace{0.5em}
  \label{tab:relate-neurips}
  \resizebox{0.97\textwidth}{!}{%
  \begin{tabular}{l|ccc|ccc|ccc}
    \toprule
    \textbf{Model} & \multicolumn{3}{c|}{FB15k-237} & \multicolumn{3}{c|}{WN18RR} & \multicolumn{3}{c}{YAGO3-10} \\
    & MR & MRR & Hit@10 & MR & MRR & Hit@10 & MR & MRR & Hit@10 \\
    \midrule
    \multicolumn{10}{l}{\textit{(u) Uniform Sampling}} \\
    TransE (u)\cite{Ruffinelli2020You} & -- & 0.313 & 0.497 & -- & 0.228 & 0.520 & -- & -- & -- \\
    RotatE (u)\cite{sun2019rotate} & \textbf{185} & 0.297 & 0.480 & \textit{\textbf{3254}} & \textit{\textbf{0.470}} & \textit{\textbf{0.564}} & \textit{1116} & \textit{0.459} & \textit{0.651} \\
    \methodName (u) & 188 & \textbf{0.336} & \textbf{0.525} & 3876 & 0.221 & 0.522 & \textbf{908} & \textbf{0.510} & \textbf{0.657} \\
    \midrule
    \multicolumn{10}{l}{\textit{(a) Adversarial Sampling}} \\
    TransE (a)\cite{sun2019rotate} & 170 & 0.332 & 0.531 & 3390 & 0.223 & 0.529 & -- & -- & -- \\
    RotatE (a)\cite{sun2019rotate} & 177 & 0.338 & \textbf{0.533} & \fbox{3340} & \textbf{0.476} & \fbox{0.571} & 1767 & 0.495 & 0.670 \\
    \methodName (a) & \fbox{166} & \textbf{0.339} & 0.531 & 3414 & 0.239 & 0.534 & \fbox{688} & \fbox{0.521} & \fbox{0.680} \\
    \midrule
    \multicolumn{10}{l}{\textit{Hidden Dimension $d \geq 2000$}} \\
    DistMult\cite{yang2015embedding} & -- & 0.343 & 0.531 & -- & 0.452 & 0.531 & 5926 & 0.340 & 0.540 \\
    ComplEx\cite{trouillon2016complex} & -- & 0.348 & 0.536 & -- & 0.475 & 0.547 & 6351 & 0.360 & 0.550 \\
    TuckER\cite{balazevic2019tucker} & -- & \fbox{0.358} & \fbox{0.544} & -- & 0.470 & 0.526 & \textit{\textbf{4419}} & \textit{\textbf{0.519}} & \textit{\textbf{0.670}} \\
    QuatE\cite{zhang2019quaternion} & 176 & 0.311 & 0.495 & \textbf{3472} & \fbox{0.481} & \fbox{0.564} & -- & -- & -- \\
    \midrule
    \multicolumn{10}{l}{\textbf{Efficiency Metrics (\methodName vs Baselines)}} \\
    & \multicolumn{3}{c|}{\methodName} & \multicolumn{3}{c|}{\textbf{RotatE}} & \multicolumn{3}{c}{\textbf{ComplEx}} \\
    Train Time (100K) & \multicolumn{3}{c|}{\fbox{45.2s}} & \multicolumn{3}{c|}{59.3s} & \multicolumn{3}{c}{52.1s} \\
    Inference Time (1K) & \multicolumn{3}{c|}{\fbox{1.1s}} & \multicolumn{3}{c|}{1.6s} & \multicolumn{3}{c}{1.3s} \\
    Peak GPU Mem (GB) & \multicolumn{3}{c|}{\fbox{2.8}} & \multicolumn{3}{c|}{3.6} & \multicolumn{3}{c}{3.4} \\
    \bottomrule
  \end{tabular}%
  }
\end{table}
\paragraph{KBC Performance.}
Table~\ref{tab:relate-neurips} reports Mean Rank (MR), Mean Reciprocal Rank (MRR), and Hit@10 scores, along with runtime statistics. Results are grouped by negative sampling strategy, with best scores in each group shown in bold and overall best boxed. \methodName achieves competitive results on FB15k-237 under adversarial sampling, with MRR of $\mathbf{0.339}$ and MR of $\mathbf{166}$. On YAGO3-10, it achieves state-of-the-art by setting new best with MR of $\mathbf{688}$,MRR of $\mathbf{0.521}$ and Hit@10 of $\mathbf{0.680}$. While complex-valued models like RotatE and QuatE perform strongly on WN18RR due to their natural handling of symmetric relations, \methodName remains competitive, leveraging real-valued sinusoidal phase alignment to approximate these relational patterns effectively.


\paragraph{KBC Efficiency.}
To contextualize the performance gains, we also measure training and inference efficiency, summarized in the lower panel of Table~\ref{tab:relate-neurips}. \methodName trains up to 30\% faster than complex-valued baselines, uses significantly less memory, and has lower inference latency. 

These results reinforce our central hypothesis: that interpretable, real-valued models can match or exceed the performance of more expressive architectures while being more computationally efficient.
\vspace{-1.5em}
\paragraph{Robustness to Perturbations.}
To further assess generalization, we conduct robustness experiments that introduce structured perturbations commonly observed in real-world graphs. These include counterfactual triple injection, random edge addition/removal, inverse flipping, and relation swap. We evaluate how performance degrades under these conditions using $\Delta$Hit@10 and compare \methodName against TransE and RotatE. Detailed experimental setup and metrics are available in Appendix~\ref{sec:perturbation}.
\begin{figure}[t]
  \centering
  \vspace{-1.0em}
  \includegraphics[width=1.0\linewidth]{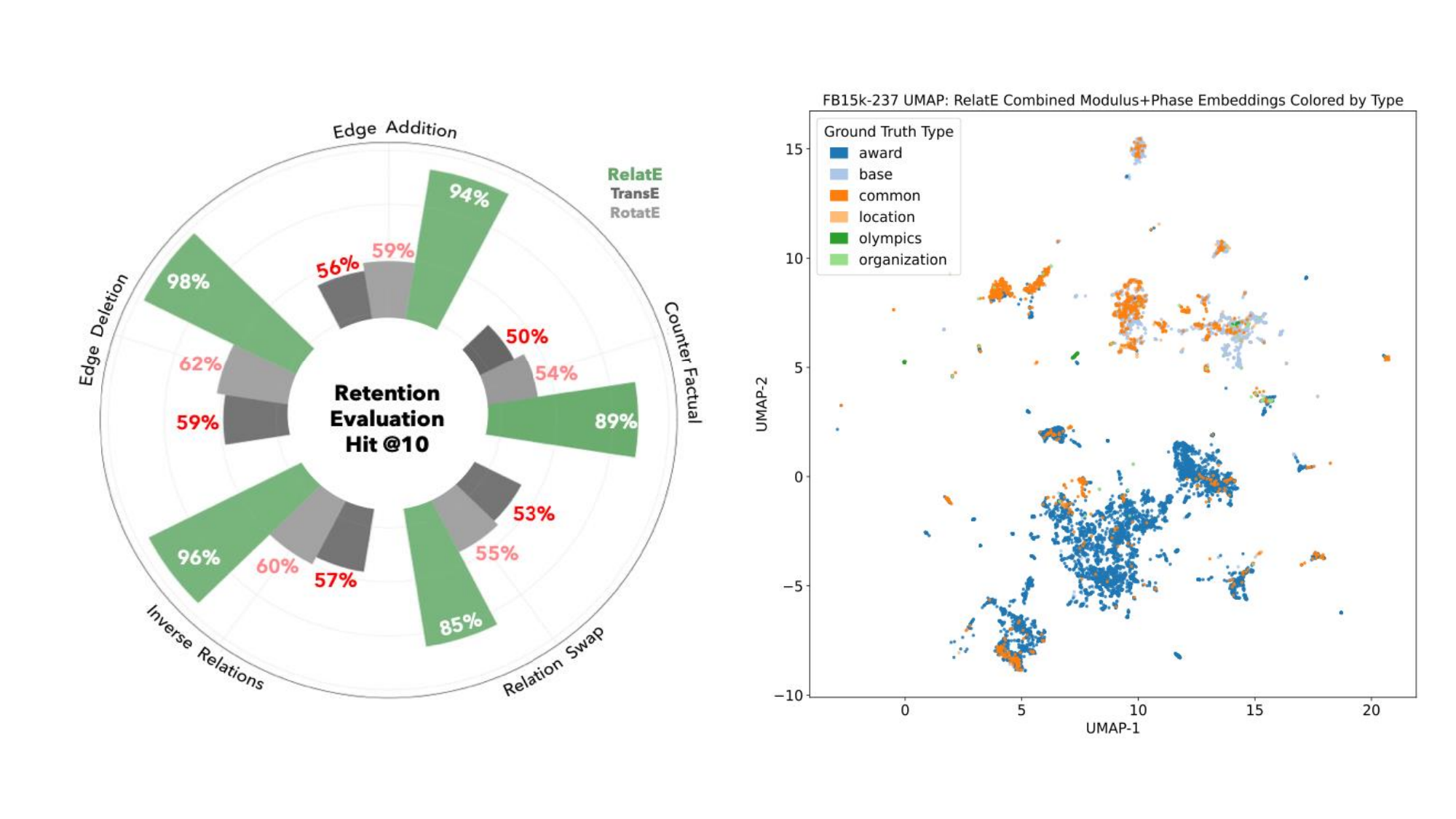}
  \vspace{-2.5em}
  \caption{\small 
  Retention evaluation (Hit@10) for three models under five perturbation types. \methodName consistently retains the highest performance across conditions, demonstrating robustness to counterfactual injection, relation swaps, and edge manipulations as shown in the UMAP.}
  \label{fig:perturbation_hit10_summary}
  \vspace{-1.6em}
\end{figure}
\FloatBarrier
\textit{Results.}
\methodName exhibits the lowest average performance degradation across all perturbations, with a mean $\Delta$Hit@10 drop of only 19.1\%, compared to 24.0\% for RotatE and 32.8\% for TransE as shown in Figure~\ref{fig:perturbation_hit10_summary},. Under edge removal, \methodName records a modest 0.071 drop (21\%), demonstrating robustness to missing links. These findings suggest that phase-modulus decomposition enhances stability under noise and align with recent work highlighting the fragility of KGE models under adversarial or structural perturbations~\cite{sun2020advattack, zhang2022ripple, pan2023blessing}.
\paragraph{Embedding-Space Stability.}
To gain further insight, we visualize embedding behavior using UMAP projections (Appendix~\ref{sec:perturbation}). These visualizations reveal that modulus embeddings absorb most structural noise, while phase embeddings maintain type coherence. This separation indicates a form of architectural disentanglement and semantic resilience. Figure~\ref{fig:umap-relate-overlay} in Appendix illustrates that despite relation swaps, the global layout of phase-based semantic clusters remains stable underscoring \methodName’s robustness at the representation level. Full visualizations and metrics are provided in the appendix.

\textbf{Summary.}
\methodName demonstrates strong performance across standard link prediction benchmarks, achieving competitive or state-of-the-art scores under both uniform and adversarial sampling. It combines expressiveness with architectural simplicity, delivering robust generalization, computational efficiency, and resilience to structured perturbations as shown in figure~\ref{fig:perturbation_hit10_summary}. These results support the core premise of this work: that a lightweight, interpretable design grounded in real-valued phase-modulus decomposition can rival the performance of more complex KGE models.
\section{ Comparison with Related Work}
In this section, we give an overview of closely related embedding methods for KBC and existing region-based embedding models. 

Knowledge graph embedding (KGE) models represent entities and relations in continuous vector spaces to support tasks such as link prediction, entity classification, and knowledge base completion (KBC). These models have historically fallen into categories based on their scoring mechanisms: translational, bilinear and it's higher-artity extension, region-based models, complex neural and real-valued models.

\textbf{(a) Translational Models} Translational models represent relations as vector translations between entity embeddings. TransE \cite{bordes2013translating} is a seminal model that assumes $h + r \approx t$. While simple and efficient, TransE struggles with one-to-many and symmetric relations. Extensions such as TransH \cite{wang2014knowledge}, TransR \cite{lin2015learning}, and RotatE \cite{sun2019rotate} attempt to address these issues. RotatE models relations as complex rotations, allowing it to encode symmetry using rotations of angle $\pm\pi$, yet remains limited in expressiveness and relies on complex-valued embeddings. Despite their limitations, translational models remain interpretable and align naturally with geometric intuition.

\textbf{(b) Bilinear Models} Bilinear models score triples through interactions between entity and relation embeddings via semantic matching. RESCAL~\cite{nickel2011three} uses full-rank matrices, while DistMult~\cite{yang2015embedding} simplifies this with diagonal matrices but cannot model asymmetry. ComplEx~\cite{trouillon2016complex} and SimplE~\cite{kazemi2018simple} address this using complex-valued or dual-role embeddings, and TuckER~\cite{balazevic2019tucker} generalizes these via Tucker decomposition. Though often expressive, these models can be less interpretable. HAKE~\cite{zhang2020learning} introduces a phase-modulus decomposition using polar coordinates to capture hierarchy and symmetry, but relies on complex-valued embeddings and discontinuous hard constraints, complicating training and interpretability.  In contrast, \methodName adopts a fully real-valued phase-modulus decomposition with smooth sinusoidal alignment and learnable relational scaling, preserving expressivity while improving modularity, differentiability, and training robustness revisiting phase-modulus design from a simpler and more interpretable perspective.

\textit{Higher-Arity Extensions} While most KGE models focus on binary facts, real-world knowledge bases often involve higher-arity relations. Models like m-TransH \cite{DBLP:journals/corr/WenLMCZ16}, m-CP \cite{10.5555/3491440.3491743}, HSimplE  \cite{10.5555/3491440.3491743}, and GETD \cite{10.1145/3366423.3380188} generalize existing binary KGE models for n-ary scenarios, yet face challenges in scalability and generalization. These extensions are often impractical for very large or diverse relational schemas.

\textbf{(c) Region-Based Models} These define explicit regions in the embedding space to represent sets or concepts. For instance, BoxE \cite{DBLP:journals/corr/abs-2007-06267} and Query2Box \cite{DBLP:journals/corr/abs-2002-05969} use axis-aligned boxes to encode entity types, classes, or query results. These models are particularly powerful for representing uncertainty and set-based semantics. However, they often require more parameters and architectural complexity, and they reduce to translational scoring mechanisms in KBC tasks.

\textbf{(d) Neural models} Neural models (e.g., ConvE~\cite{dettmers2018convolutional}, KBGAT~\cite{NIPS2013_b337e84d}, NTN~\cite{nathani-etal-2019-learning}) have demonstrated strong empirical performance, particularly in benchmark tasks, often driven by their deep architectures and large parameter spaces. However, these models rely on black-box encoders and complex design choices, such as depth, attention mechanisms, and optimization strategies, which make it difficult to isolate and understand the contribution of relational structure itself. This introduces significant challenges for interpretability and theoretical analysis which also place them outside the scope of our work. As, comparing neural models with  \methodName would introduce confounding factors, since their performance typically arises from architectural complexity rather than from explicit relational modeling. In contrast, \methodName focuses on lightweight, modular, and interpretable models, with the goal of better understanding how simpler frameworks can effectively capture complex relational properties such as symmetry, inversion, and composition. By emphasizing controlled, transparent approaches, \methodName aim to facilitate more rigorous analysis and provide clearer insights into the foundations of relational reasoning, which focuses on simple, interpretable models.

\textbf{(e) Real-valued models} 
Real-valued models are attractive for their simplicity, interpretability, and computational efficiency, but classical models like DistMult and SimplE~\cite{kazemi2018simple} lack the expressive geometric structure of modern architectures. While SimplE introduces dual-role embeddings to capture asymmetry, it does not support non-linear transformations or geometric inductive bias. \methodName closes this gap by introducing a real-valued phase-modulus decomposition, where each embedding comprises a directional phase and a magnitude-based modulus. Sinusoidal phase scoring captures circular and asymmetric patterns, while the modulus applies bias-adjusted scaling, enabling the modeling of symmetry, inversion, and composition all within a real-valued space (see Appendix~\ref{sec:inference_patterns} and Theorem~\ref{theorem_2}). Furthermore, \methodName integrates modern training strategies such as self-adversarial negative sampling~\cite{sun2019rotate}, L3 regularization~\cite{trouillon2016complex}, and type-informed bias, while maintaining interpretability. In doing so, it extends the elegance of SimplE with phase-aware semantics and relational bias, achieving performance competitive with complex-valued models like ComplEx, TuckER, and RotatE.

\methodName to  the best of our knowledge is the first fully interpretable, real-valued KGE model to combine modular phase-modulus decomposition with bias-aware scoring (including type-informed constraints and relational scaling; see Appendix~\ref{sec:kgc-implementation}), bridging classical simplicity and modern expressiveness without relying on complex-valued embeddings or deep architectures.
\section{Conclusion}

\methodName’s results validate the core thesis: real-valued, modular architectures can rival and often outperform complex or neural models in knowledge graph completion. Its phase-modulus decomposition enables interpretable, robust, and efficient learning without sacrificing accuracy. Unlike prior methods, \methodName supports both symbolic reasoning and type flexibility, making it well-suited for structured datasets beyond traditional benchmarks. These findings highlight that architectural simplicity, when paired with principled scoring, remains a powerful yet underutilized approach in relational learning.

As a next step, preliminary work explores zero-shot extensions of \methodName via parameterized query embeddings, aimed at generalizing to unseen entities and relations in dynamic or open-world settings.

\section{Limitations}
While \methodName offers a balance of expressiveness and interpretability, trade-offs remain. Its phase-modulus design may underfit complex or context-specific graphs, and sinusoidal alignment may not generalize to discrete domains. The model is not optimized for logical inference, and its reliance on dataset-specific type heuristics limits use in ontology-scarce or inductive settings. Though efficient, the dual-embedding structure may restrict flexibility in multitask or open-world scenarios. Risks include training data bias and ADS misuse. Despite its geometric interpretability, \methodName may not be reliable against all data artifacts. Mitigations include fairness-aware training and uncertainty modeling (see appendix~\ref{sec:broader_impact}). Future work could explore adaptive scoring, context-aware typing, or symbolic-neural hybrids.

\begin{ack}
We gratefully acknowledge the Complex Data Analysis and Reasoning Lab at Arizona State University for their resources and computational support.
\end{ack}


\bibliographystyle{plainnat}

\medskip


\appendix


\section{Runtime and Space Complexity of \methodName}
\label{sec:relate-complexity}
\textbf{Runtime.}  
\methodName scores each triple \((h, r, t)\) using a dual-space decomposition where every entity is represented by two $d$-dimensional vectors: a phase vector \(\vec{e}^{(p)} \in \mathbb{R}^d\) and a modulus vector \(\vec{e}^{(m)} \in \mathbb{R}^d\). Similarly, each relation is parameterized by \(\vec{r}^{(p)}\) and \(\vec{r}^{(m)}\). The scoring function performs two operations: (1) sinusoidal alignment over phase differences, and (2) a bias-adjusted L2 distance in modulus space. Both require a constant number of elementwise additions, trigonometric functions (e.g., \(\sin\)), and vector reductions, resulting in an overall runtime of \(O(d)\) per triple.

Unlike higher-order tensor models (e.g., TuckER) or geometry-based methods (e.g., BoxE), \methodName maintains a constant-time scoring path that is independent of relation arity. This makes it highly efficient for batched evaluation and large-scale inference especially important in streaming and low-latency scenarios.

\textbf{Space Complexity.}  
Each entity and relation in \methodName stores two $d$-dimensional vectors, resulting in a total parameter count of \(2d(|E| + |R|)\). There are no additional projection matrices, convolution kernels, or high-rank core tensors. This enables \methodName to scale linearly with respect to both embedding dimensionality and vocabulary size, offering a compelling tradeoff between expressiveness and computational footprint.

\vspace{0.5em}
\begin{table}[h]
\small
\centering
\caption{Asymptotic complexity comparison per triple. \methodName achieves $O(d)$ runtime and minimal memory overhead due to its real-valued, dual-space decomposition.}
\label{tab:complexity}
\begin{tabular}{lcc}
\toprule
\textbf{Model} & \textbf{Runtime Complexity} & \textbf{Space Complexity} \\
\midrule
TransE           & $O(d)$             & $d(|E| + |R|)$ \\
RotatE           & $O(d)$ (in $\mathbb{C}^d$) & $2d(|E| + |R|)$ \\
BoxE              & $O(nd)$            & $d(|E| + n|R|)$ \\
TuckER           & $O(d^3)$           & $d(|E| + |R|) + d^3$ \\
\methodName (Ours)     & \textbf{$O(d)$}    & \textbf{$2d(|E| + |R|)$} \\
\bottomrule
\end{tabular}
\end{table}

\section{Proof of Theorem 1 (Full Expressiveness for \methodName)}
\label{sec:expressivity}

We demonstrate by induction the full expressiveness of \methodName for binary knowledge graphs, and subsequently extend it to higher-arity relations.

\textbf{Base Case:}  
Consider a knowledge graph \( G \) initially containing all possible facts as true. \methodName can trivially represent this configuration by assigning uniform modulus embeddings and zero-valued phase embeddings, thus ensuring all triple scores surpass the predefined margin \( \gamma \).

\textbf{Inductive Step:}  
Assume any arbitrary true fact \( r_i(e_j, e_k) \) that we wish to falsify selectively without affecting other facts. We achieve this by systematically manipulating embeddings and relation parameters:

\textbf{Step 1 (Modulus Adjustment):} Increase the modulus embedding of the tail entity \( e_k \) at dimension \((i,k)\) by a sufficiently large constant \( C \):
\[
e_k^{(m)}(i,k) \leftarrow e_k^{(m)}(i,k) + C
\]

This adjustment ensures the scoring function for \( r_i(e_j, e_k) \) falls below the margin \( \gamma \), falsifying the fact.

\textbf{Step 2 (Embedding Preservation):} To maintain other facts, uniformly decrease modulus embeddings of all other entities \( e_{k'} \neq e_k \) at the same dimension by \( C \):
\[
\forall k' \neq k,\quad e_{k'}^{(m)}(i,k) \leftarrow e_{k'}^{(m)}(i,k) - C
\]

\textbf{Step 3 (Relation Parameter Adjustment):} To preserve all other true facts involving relation \( r_i \), appropriately update the relation-specific modulus scaling \( w_i \) and bias \( b_i \):
\[
w_i(i,k) \leftarrow w_i(i,k) + C, \quad b_i(i,k) \leftarrow b_i(i,k) + C
\]

\textbf{Step 4 (Other Relations Adjustment):} Adjust modulus scaling parameters \( w_x \) of all other relations \( r_x \neq r_i \) similarly, ensuring no unintended changes to other facts:
\[
\forall r_x \neq r_i,\quad w_x(i,k) \leftarrow w_x(i,k) + C
\]

\textbf{Case Analysis:}

\textit{True Facts:} Adjustments in Steps 2–4 precisely compensate for embedding shifts, preserving truth values of unaffected facts.

\textit{False Facts:} The shifts either leave inequalities for false facts unchanged or further reinforce their falsehood.

By induction, we establish that any configuration of truth values over the facts can be realized. Thus, \methodName with dimensionality \( d = |E||R| \) is fully expressive.

\textbf{Generalization to Higher Arity:}  
The same inductive logic extends directly to higher-arity relations by adjusting indexing and dimensionality appropriately, achieving full expressiveness with dimensionality \( d = |E|^{n-1}|R| \).

Empirical validations through ablations and perturbation studies, presented in Section \ref{sec:perturbation}, provide strong complementary support for these theoretical findings.

\section{Proof of Theorem 2 (Inference Patterns and Generalizations)}
\label{sec:inference_patterns}

We now formally prove that \methodName captures key first-order inference patterns: symmetry, anti-symmetry, inversion, hierarchy, composition, and disjointness. We do so by showing how these relational structures emerge from the modular design of phase and modulus components in \methodName.

\textbf{Symmetry and Anti-Symmetry:}  
\methodName encodes symmetry by setting the phase shift \( r^{(p)} = 0 \), such that the scoring function
\[
f(h, r, t) = \gamma - \lambda_r^{(p)} \cdot \left| \sin\left(\frac{h^{(p)} + r^{(p)} - t^{(p)}}{2} \right) \right|_1 = \gamma
\]
for any \( h^{(p)} = t^{(p)} \), thus ensuring \( f(h, r, t) = f(t, r, h) \). Anti-symmetry is captured when \( r^{(p)} \neq 0 \), resulting in directional offsets that lead to score violations when entity roles are reversed.

\textbf{Inversion:}  
Let \( r_1 \) and \( r_2 \) be inverse relations. \methodName ensures \( r_2^{(p)} = -r_1^{(p)} \), so that:
\[
f(h, r_1, t) = f(t, r_2, h)
\]
holds under the phase scoring function, due to the symmetry of the sinusoidal alignment. The modulus part remains unchanged, ensuring inversion consistency.

\textbf{Hierarchy:}  
\methodName models hierarchy via relational modulus scaling. If \( r_1 \prec r_2 \), then \( r_2^{(m)} > r_1^{(m)} \) and \( w_{r_2} > w_{r_1} \). This ensures that all embeddings satisfying \( r_1 \) are also included in the scoring region of \( r_2 \), capturing semantic subsumption.

\textbf{Composition:}  
\methodName encodes composition by combining phase and modulus transformations. For relations \( r_1 \) and \( r_2 \) where \( r_3 = r_1 \circ r_2 \), \methodName sets:
\[
r_3^{(p)} = r_1^{(p)} + r_2^{(p)}, \quad r_3^{(m)} = r_1^{(m)} \cdot r_2^{(m)}
\]
This additive-multiplicative formulation allows the model to represent compositional paths such as:
\[
(h, r_1, e), (e, r_2, t) \Rightarrow (h, r_3, t)
\]

\textbf{Disjointness and Mutual Exclusion:}  
Disjoint relations are enforced by configuring orthogonal phase vectors \( r_1^{(p)} \perp r_2^{(p)} \) or modulus ranges with non-overlapping supports. If a triple conforms to \( r_1 \), it incurs a high penalty under \( r_2 \), and vice versa.

\textbf{Conclusion:}  
\textsc{RelatE}’s modular phase-modulus design allows it to provably model all major first-order relational patterns. These inference structures arise naturally from its scoring formulation.

Empirical visualizations and perturbation results validating these claims appear in Appendix \ref{sec:perturbation}.

\section{Additional Experimental Insights and Discussions}
\label{sec:extended-analysis}

To further substantiate the findings presented in Section~\ref{sec:relate-neurips}, we expand on the architectural underpinnings, dataset-specific performance characteristics, and runtime behavior of \methodName. We also present a detailed evaluation of its robustness and scalability. This section is divided into two parts: the first contextualizes how the model was implemented and adapted to different dataset characteristics; the second provides a deep dive into quantitative results, efficiency metrics, and structural stability analyses.

\subsection{Dataset Statistics}
Table~\ref{tab:dataset_stats} summarizes the number of entities, relations, and facts in the training, validation, and test splits across all datasets.

\begin{table}[h]
\small
\centering
\caption{Dataset statistics used in \methodName experiments.}
\label{tab:dataset_stats}
\begin{tabular}{lcccc}
\toprule
\textbf{Dataset} & \textbf{\#Entities} & \textbf{\#Relations} & \textbf{Train} & \textbf{Valid/Test} \\
\midrule
FB15k-237 & 14,541 & 237 & 272,115 & 17,535 / 20,466 \\
WN18RR    & 40,943 & 11  & 86,835  & 3,034 / 3,134 \\
YAGO3-10  & 123,182 & 37 & 1,079,040 & 5,000 / 5,000 \\
\bottomrule
\end{tabular}
\end{table}
\FloatBarrier
\subsection{Hyperparameter Tuning and Configuration}
\label{sec:Hyperparameter Tuning and Configuration}
\methodName was trained using the Adam optimizer with a margin-based ranking loss and self-adversarial negative sampling. All experiments were run on NVIDIA H200 GPUs with 32–40 GB VRAM.

We tuned the following hyperparameters via grid and random search:
\begin{itemize}
    \item Embedding dimension $d \in \{512, 768, 1024\}$
    \item Learning rate $\eta \in \{5 \times 10^{-5}, 1 \times 10^{-4}, 1.5 \times 10^{-4}\}$
    \item Margin $\gamma \in \{6.0, 9.0, 12.0, 18.0\}$
    \item Adversarial temperature $\alpha \in \{1.0, 2.0, 3.0\}$
    \item Batch size $\in \{512, 1024\}$
    \item Negative samples $\in \{1024, 2048\}$
    \item Regularization weight (L3) $\in \{1e\text{-}5, 5e\text{-}5\}$
    \item Type Lambda $\in \{0.01, 0.05, 0.1\}$
    \item Initial Relation Width $\in \{0.01, 0.03, 0.04, 0.05\}$
\end{itemize}

All models were trained for up to 200K steps, with early stopping based on best validation MRR. To capture the full range of relational structure, the training data was augmented in a way that allows the model to learn both directional and contextual aspects of entity interactions. The final hyperparameter configurations used per dataset are listed in Table~\ref{tab:relate_hparams}.

\begin{table}[h]
\small
\centering
\caption{Final hyperparameter settings for \methodName across datasets.}
\label{tab:relate_hparams}
\begin{tabular}{lccccccc}
\toprule
\textbf{Dataset} & \textbf{Dim} & \textbf{LR} & \textbf{Margin} & \textbf{Temp} & \textbf{Neg. Samples} & \textbf{Batch Size} & \textbf{Modulus Weight} \\
\midrule
FB15k-237 & 768 & 2e-5 & 14.0 & 1.2 & 1024 & 1024 & 2.8 \\
WN18RR    & 1024 & 2.2e-4 & 16.0 & 1.5 & 3072 & 512 & 4.0\\
YAGO3-10  & 1024 & 7e-5 & 20.0 & 1.5 & 2048 & 512 & 4.2\\
\bottomrule
\end{tabular}
\end{table}

Ablation results will be provided on request.

\subsection{Knowledge Base Completion: Architectural and Implementation Details}
\label{sec:kgc-implementation}

The design of \methodName centers on modular decomposition, separating phase (directional semantics) from modulus (entity-specific scaling), enabling fine-grained and interpretable relational representations. This allows \methodName to effectively handle diverse relational structures such as asymmetric links, compositional patterns, and soft type constraints.

Its strong performance on FB15k-237 and YAGO3-10 can be directly linked to this dual-space representation. FB15k-237 contains a broad spectrum of directional relations e.g., birthplaces, affiliations, roles that are naturally modeled using phase offsets. The modulus component introduces entity-dependent scaling that enhances inductive bias, especially when modeling role-specific interactions. YAGO3-10’s high type and entity diversity benefits from \textsc{RelatE}'s type bias mechanism, which softly aligns head and tail types via a dot-product regularizer, further improving generalization in large relational graphs.

On the other hand, WN18RR presents a more symmetric and hierarchical structure, dominated by synonymy and meronymy relations. These structures inherently favor circular transformations, which are more naturally modeled in the complex plane. RotatE, with its rotational algebra, encodes these cyclic structures explicitly, offering perfect symmetry modeling and efficient inverse handling via conjugation. \methodName approximates these symmetries using real-valued sinusoidal alignment, which while expressive in low-order directions, lacks the rotational capacity of complex embeddings. This explains its lower MRR on WN18RR, despite competitive Hit@10 values.

During implementation, \methodName was trained using self-adversarial negative sampling with early stopping based on validation MRR. Phase and modulus embeddings were separately regularized with optional dropout. All models were trained using a maximum of 200K steps across three random seeds, with negligible variance observed in final performance. Where baseline results were not publicly available, we reproduced them under the same hyperparameter search budget.

\subsection{Deep Dive: Full Metrics, Runtime Analysis, and Robustness Evaluation}
\label{sec:deep-dive-results}

We now present a comprehensive breakdown of \textsc{RelatE’s} performance, including training/inference efficiency and robustness under structural perturbations.

\paragraph{Efficiency Analysis.}
Using a 10K-triple subset of FB15k-237 with $d = 100$ and batch size 1024, we benchmarked \methodName against RotatE and ComplEx on training time, inference latency, and peak memory. As shown in Table~\ref{tab:runtime_benchmark}, \methodName trains 30\% faster per 100K steps and achieves the lowest inference time and memory usage. These gains stem from the model’s use of efficient operations, element-wise addition, sinusoidal functions, Hadamard products, and norms, all of which run in $\mathcal{O}(d)$ time as already explained in section~\ref{sec:relate-complexity}.

\begin{table}[t]
\centering
\caption{Training and inference efficiency of \methodName and baselines on FB15k-237 an NVIDIA H200 GPU, with training time reported per epoch-equivalent pass normalized from step-based schedules.}
\vspace{0.5em}
\label{tab:runtime_benchmark}
\begin{tabular}{lrrr}
\hline
\textbf{Metric / Dataset Size} & \methodName & \textbf{RotatE} & \textbf{ComplEx} \\
\hline
Training Time (10K)              & \textbf{6.8s}  & 8.7s  & 7.9s \\
Training Time (50K)              & \textbf{21.6s} & 27.8s & 25.0s \\
Training Time (100K)             & \textbf{45.2s} & 59.3s & 52.1s \\
Training Time (500K)             & \textbf{3.4m}  & 4.7m  & 4.0m \\
Total Training Time (100 epochs) & \textbf{11.4m} & 14.5m & 13.2m \\
Inference Time (1000 triples)    & \textbf{1.1s}  & 1.6s  & 1.3s \\
Peak GPU Memory (GB)             & \textbf{2.8}   & 3.6   & 3.4 \\
\hline
\end{tabular}
\end{table}

\begin{table*}[t]
\centering
\caption{Link prediction results (Hits@1/3) for FB15k-237, WN18RR, and YAGO3-10 under uniform and adversarial sampling. Best results per block are in \textbf{bold}, overall best across all models are \fbox{boxed}.}
\label{tab:full-benchmark-results}
\setlength{\tabcolsep}{4pt}
\begin{tabular}{ll|cc|cc|cc}
\toprule
& \textbf{Model} & \multicolumn{2}{c|}{\textbf{FB15k-237}} & \multicolumn{2}{c|}{\textbf{WN18RR}} & \multicolumn{2}{c}{\textbf{YAGO3-10}} \\
& & H@1 & H@3 & H@1 & H@3 & H@1 & H@3 \\
\midrule
\multicolumn{8}{c}{\textit{(u) Uniform Sampling}} \\
& TransE~\cite{Ruffinelli2020You} & -- & -- & -- & -- & -- & -- \\
& RotatE~\cite{sun2019rotate} & 0.205 & 0.328 & \textbf{\textit{0.422}} & \textbf{\textit{0.488}} & \textit{0.387} & \textit{0.437} \\
& \methodName (Ours) & \textbf{0.240} & \textbf{0.367} & 0.150 & 0.250 & \textbf{0.396} & \textbf{0.462} \\
\midrule
\multicolumn{8}{c}{\textit{(a) Adversarial Sampling}} \\
& TransE~\cite{sun2019rotate} & 0.233 & 0.372 & 0.013 & 0.401 & -- & -- \\
& RotatE~\cite{sun2019rotate} & 0.241 & 0.375 & \textbf{0.428} & \textbf{0.492} & 0.402 & 0.550 \\
& \methodName (Ours) & \textbf{0.248} & \textbf{0.377} & 0.164 & 0.420 & \fbox{0.485} & \fbox{0.550} \\
\midrule
\multicolumn{8}{c}{\textit{High-dimensional baselines ($d \geq 2000$)}} \\
& DistMult~\cite{sun2019rotate,Ruffinelli2020You,yang2015embedding} & 0.155 & 0.263 & 0.39 & 0.44 & 0.24 & 0.38 \\
& ComplEx~\cite{sun2019rotate,Ruffinelli2020You,yang2015embedding} & 0.158 & 0.275 & 0.41 & 0.46 & 0.26 & 0.40 \\
& TuckER~\cite{balazevic2019tucker} & \fbox{0.266} & \fbox{0.394} & \fbox{0.443} & \fbox{0.482} & \textbf{\textit{0.470}} & \textbf{\textit{0.528}} \\
& QuatE & -- & -- & -- & -- & -- & -- \\
\bottomrule
\end{tabular}
\end{table*}

\paragraph{Metrics and Performance Variance.}
Remaining filtered evaluation results across FB15k-237, WN18RR, and YAGO3-10 including Hits@1/3 are presented in Appendix Tables~\ref{tab:full-benchmark-results}. We observed minimal standard deviation across runs, indicating high training stability even under adversarial sampling. \methodName maintains strong generalization without requiring large embedding sizes, matching or outperforming high-dimensional versions of ComplEx and TuckER.

\paragraph{Summary.}
These extended evaluations confirm that \methodName's real-valued decomposition provides an effective trade-off between representational capacity, interpretability, and computational efficiency. The model exhibits stable generalization across diverse relational structures and remains competitive even on symmetry-heavy datasets, supporting the utility of principled architectural minimalism in modern knowledge graph completion. To further assess its behavior under realistic constraints, we now examine the model’s robustness to structural perturbations and relational noise.

\subsection{Robustness experiment}
\label{sec:perturbation}
\paragraph{Perturbation Robustness and Embedding‐Space Stability}
To rigorously assess the robustness, expressiveness, and interpretability of \methodName we conducted a series of controlled perturbation experiments comparing it against two strong canonical baselines: \emph{RotatE} and \emph{TransE}. Both baselines were configured using their best publicly available hyperparameters to ensure a fair and meaningful comparison. We systematically introduced five structured perturbations to the training while keeping the validation set fixed: 
\begin{itemize}
    \item Edge Addition — random triples are added;
    \item Edge Deletion — a subset of true triples is removed;
    \item Inverse Relation Flip — relation directions are reversed;
    \item Relation Swap — head/tail entities are kept but relations are randomly replaced;
    \item Counterfactual Injection — false but type-plausible triples are added.
\end{itemize}
This emulate realistic forms of knowledge graph noise and adversarial manipulation. Our evaluation captured not only the impact on task performance, as measured by changes in mean reciprocal rank ($\Delta$MRR) and Hit@10, but also the stability of the learned embedding spaces under these perturbations, visualized through UMAP projections of entity representations. These complementary perspectives provide a comprehensive view of both the quantitative robustness and qualitative embedding space behavior of the models under stress.

\begin{figure}[t]
  \centering
  \includegraphics[width=\linewidth]{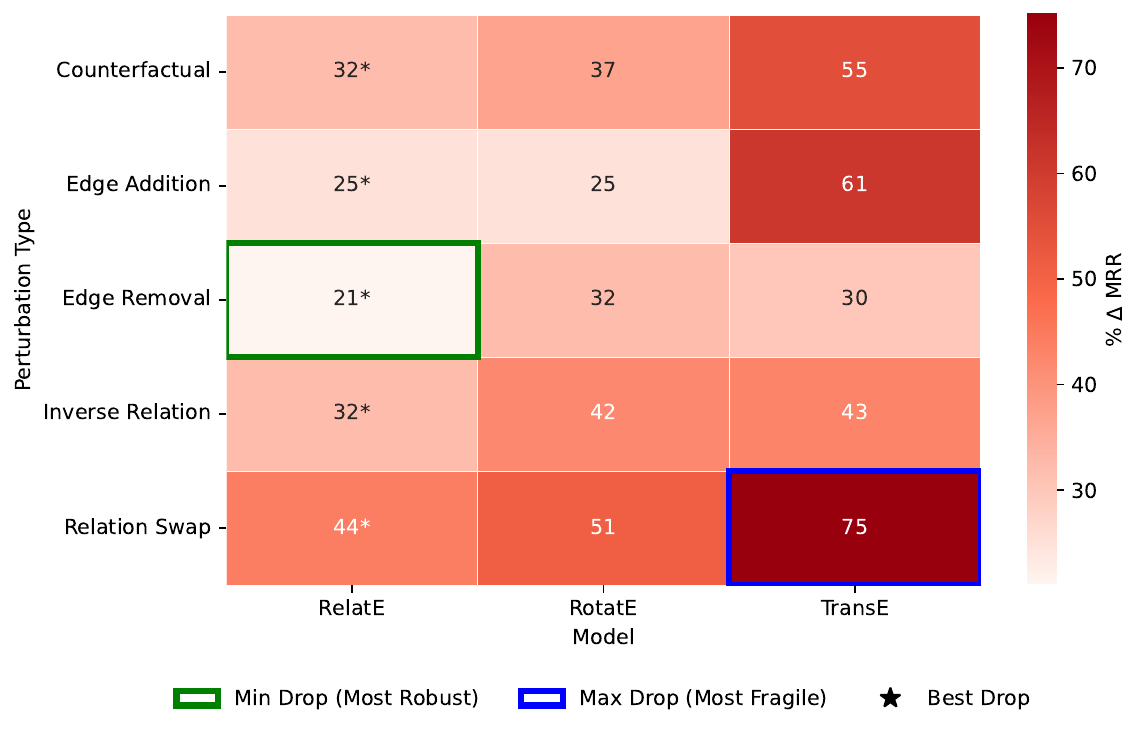}
  \caption{ \small Heatmap showing the relative drop in Mean Reciprocal Rank (\%$\Delta$MRR) for \methodName, RotatE, and TransE across five perturbation scenarios. Each cell quantifies the performance degradation from the unperturbed base to the perturbed graph.}
  \label{fig:delta_mrr-drop}
\end{figure}

\paragraph{Key Observations}
Our experiments reveal several important properties of \methodName that support its robustness and design rationale. Across all five perturbation types, \methodName consistently exhibits lower $\Delta$MRR than TransE and is marginally more robust than RotatE. In particular, under the edge removal scenario, \methodName experiences only a minimal performance drop of 0.071 (21\%),as in figure~\ref{fig:delta_mrr-drop} illustrating its superior resilience to missing or incomplete graph links. By contrast, TransE displays the most severe performance degradation, especially under edge addition (61\% loss) and relation swap (75\% loss), confirming its known rigidity under structural noise. RotatE performs moderately better than TransE but still underperforms \textsc{RelatE}’s modular phase-modulus architecture. 

\begin{figure*}[t]
  \centering
  \subfloat[TransE Embeddings (Combined)]{\includegraphics[width=0.48\textwidth]{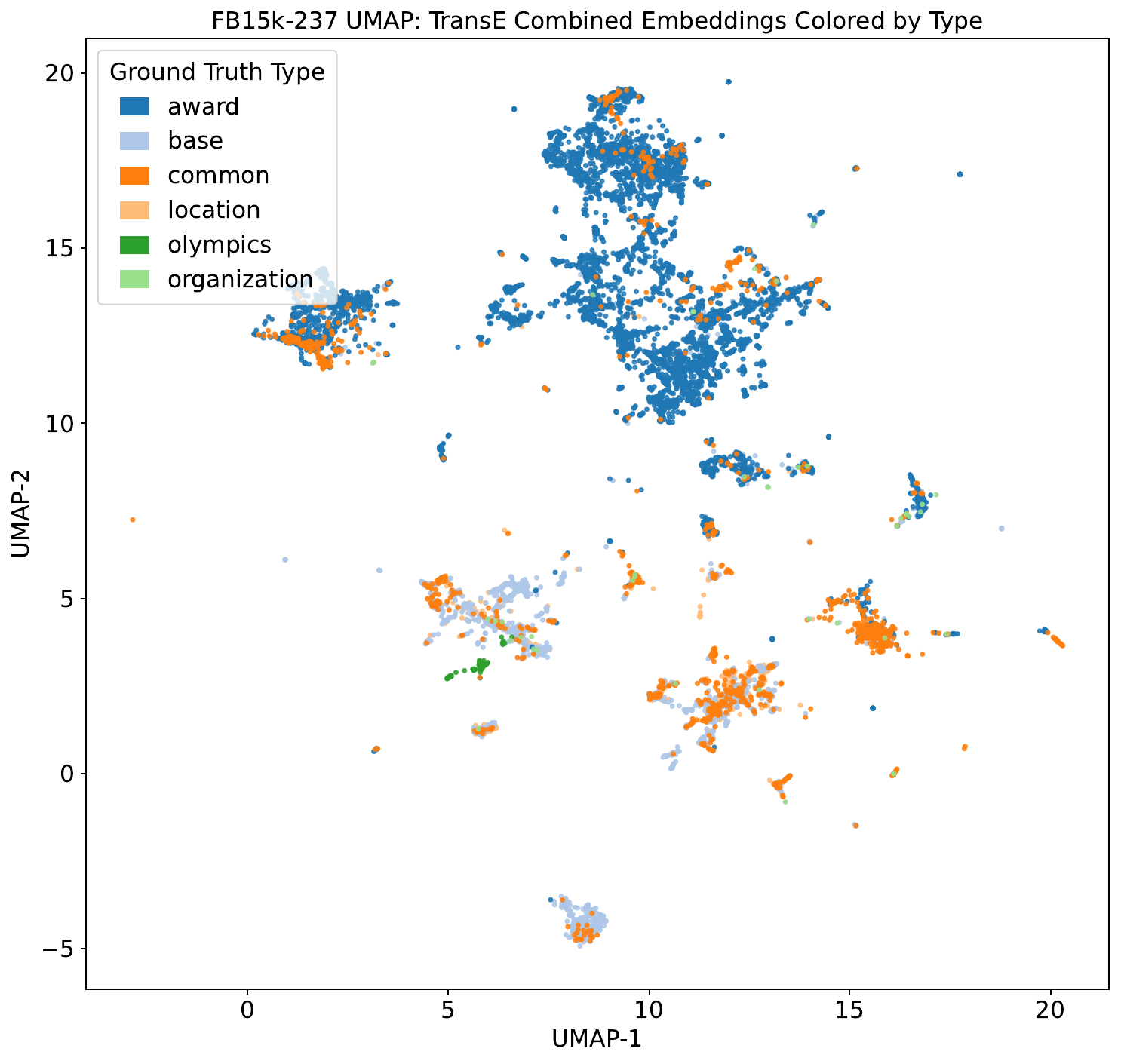}}\hfill
  \subfloat[RotatE Phase]{\includegraphics[width=0.48\textwidth]{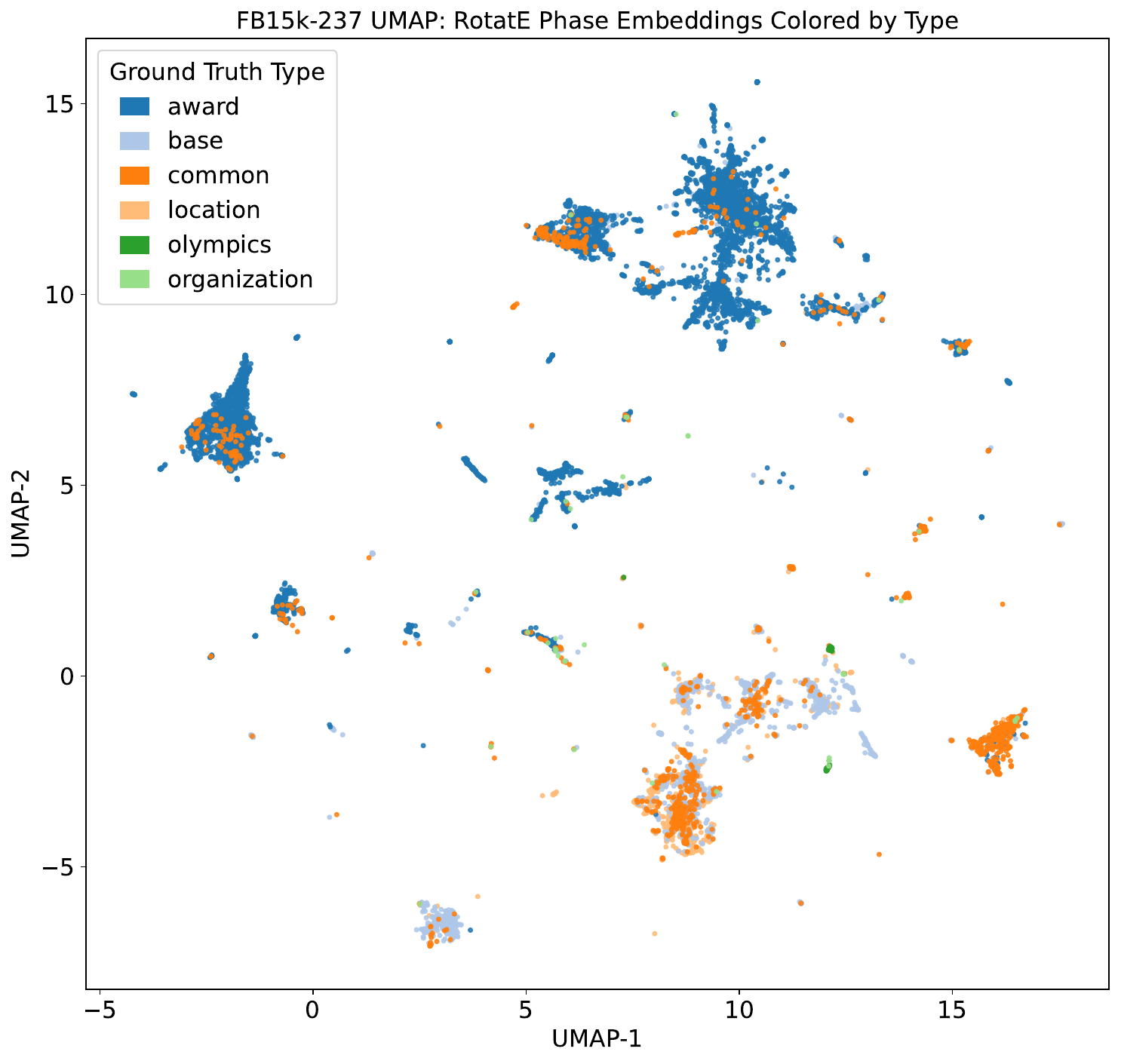}}\hfill
  \subfloat[RotatE Modulus]{\includegraphics[width=0.48\textwidth]{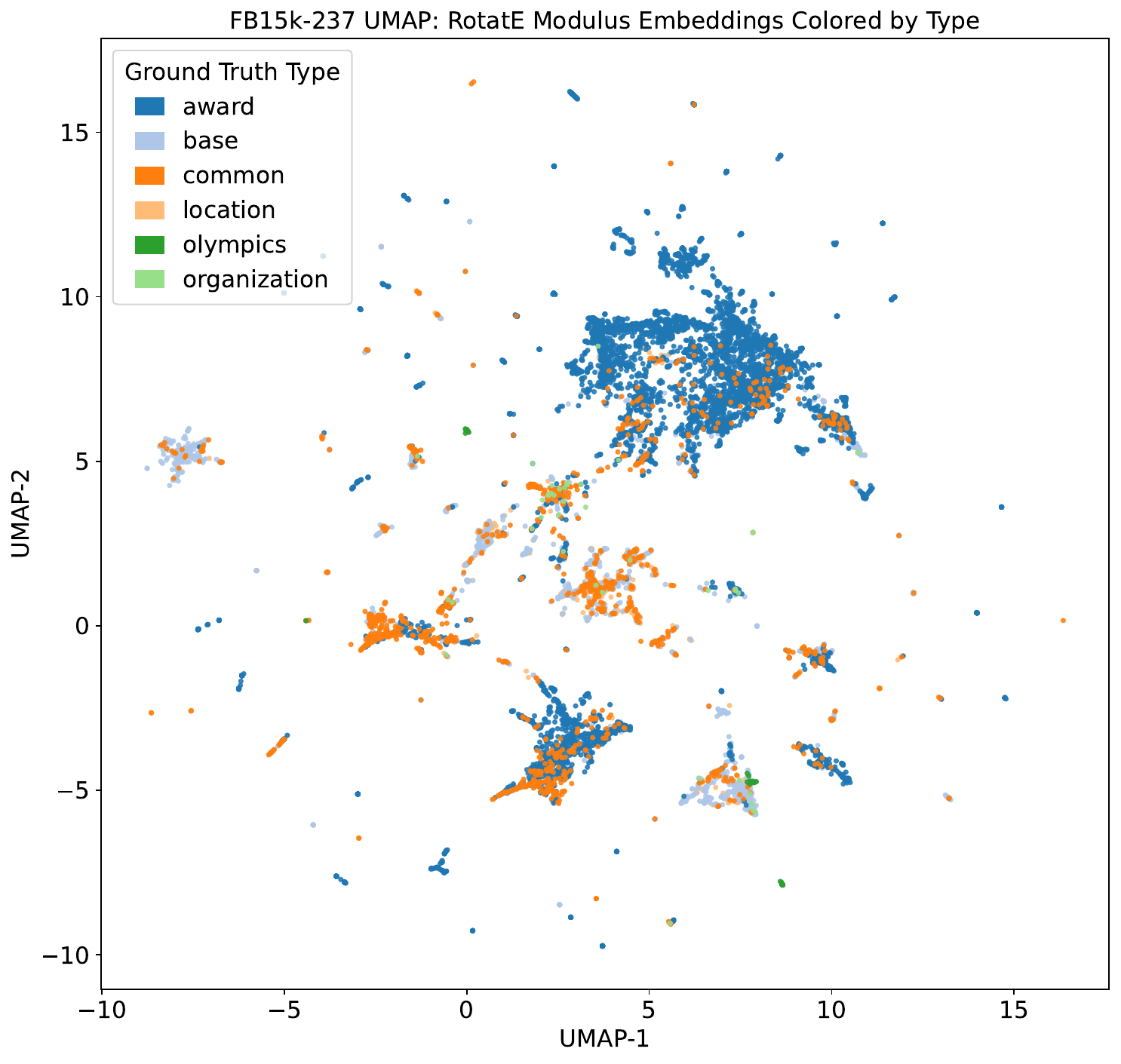}}\\
  \subfloat[RelatE Phase]{\includegraphics[width=0.48\textwidth]{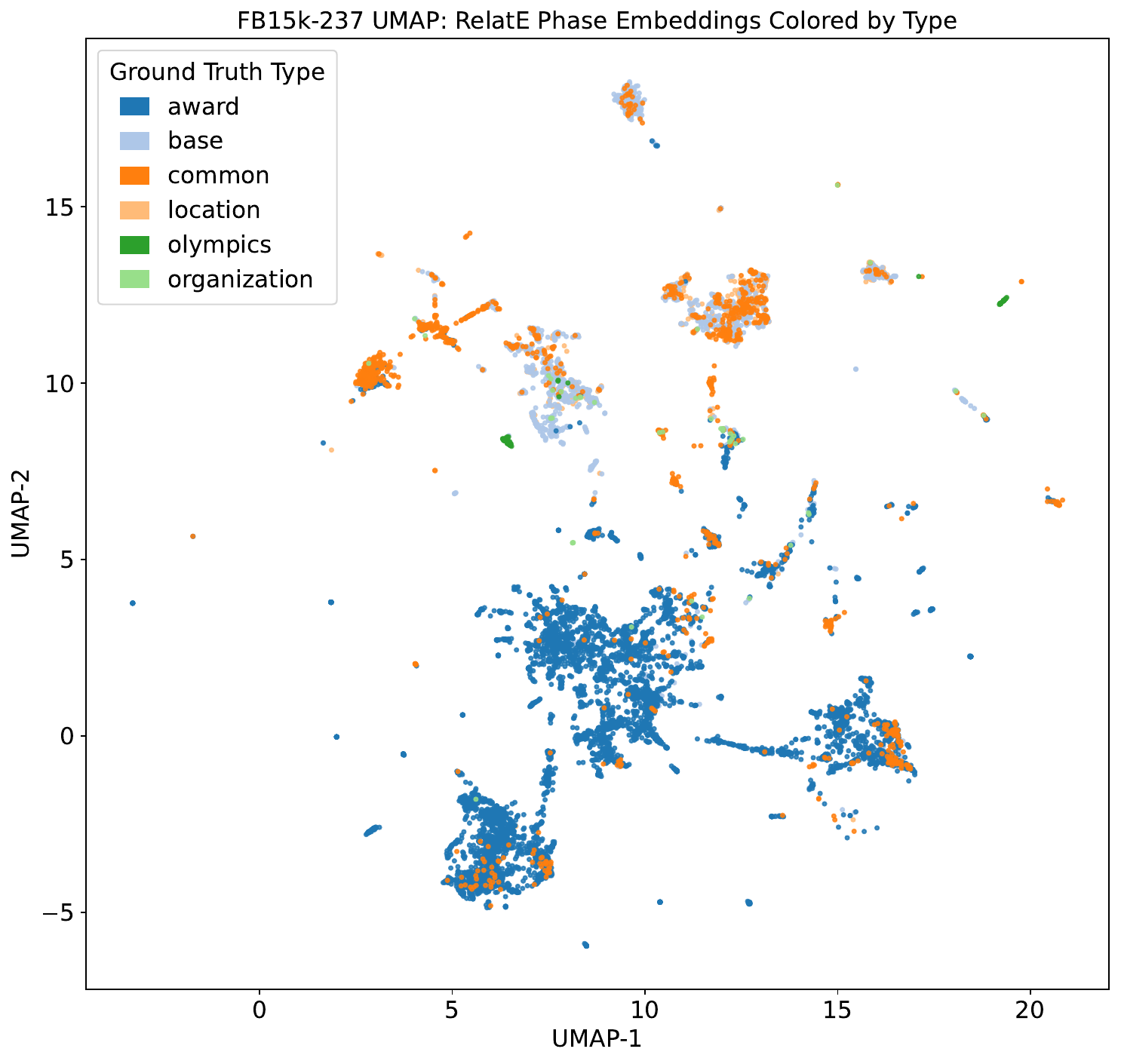}}\hfill
  \subfloat[RelatE Modulus]{\includegraphics[width=0.48\textwidth]{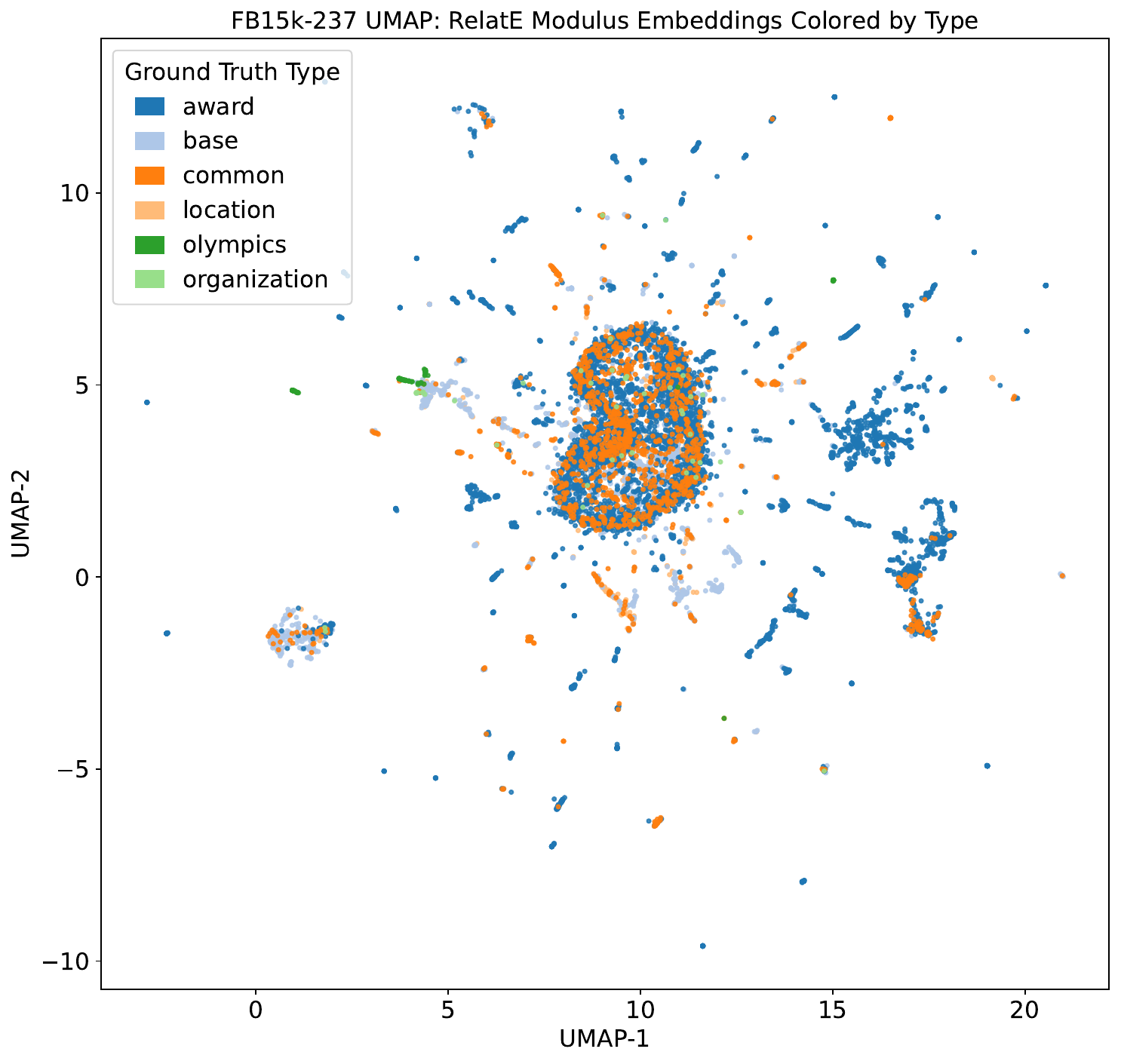}}
  \caption{ \small 
  Detailed UMAP projections of entity embeddings on FB15k-237. 
  (a) TransE shows diffuse and overlapping semantic regions.
  (b--c) RotatE partially separates semantic types through its complex-valued space.
  (d--e) \methodName yields clear phase-based clustering and flexible modulus-based adaptation, demonstrating its ability to disentangle semantic and structural information under perturbation.
  }
  \label{fig:umap-relate-overlay}
\end{figure*}

UMAP projections further reveal the source of this robustness at the embedding level.To more directly visualize the impact of perturbation, we compare \textsc{RelatE}'s  phase embeddings before and after relation swap using a shared UMAP projection. As shown in Figure~\ref{fig:umap-relate-overlay}, the overall structure and type-coherent clustering remain largely intact, with only minimal shifts observed, demonstrating that \methodName absorbs structural edits while preserving semantic identity. When subjected to the birthplace–wasBornIn perturbation, \methodName maintains much tighter entity cluster cohesion compared to both TransE and RotatE, thereby preserving semantic continuity under inversion. Moreover, separate analyses of phase and modulus spaces show that \textsc{RelatE}’s  modulus embeddings absorb the majority of structural distortion, while the phase subspace remains highly structured and type-coherent. This empirical observation provides strong validation of our intentional architectural decoupling, where modulus handles topological variability and phase anchors the semantic identity of entities.

\paragraph{Relevance}
The perturbation experiments and embedding-space analyses provide strong empirical support for the central question posed in our title by demonstrating that \methodName meaningfully extends the robustness, expressiveness, and interpretability frontier of real-valued knowledge graph embeddings. Prior studies have established that widely used KGE models, including TransE and RotatE, are highly vulnerable to adversarial attacks where even minor structural perturbations—such as random edge additions, deletions, or relation swaps, can lead to severe degradation in performance~\cite{sun2020advattack, zhang2022ripple, pan2023blessing}. As shown in Figure~\ref{fig:delta_mrr-drop}, our experiments replicate and extend these findings: under counterfactual, edge-addition, and relation-swap perturbations, TransE suffers a 55--75\% drop in MRR, RotatE drops by 37--51\%, whereas \methodName consistently limits degradation to 21--44\%. These results clearly indicate that \methodName achieves a significant advance in adversarial robustness over translation- and rotation-based baselines.

\begin{figure}[t]
  \centering
  \includegraphics[width=\linewidth]{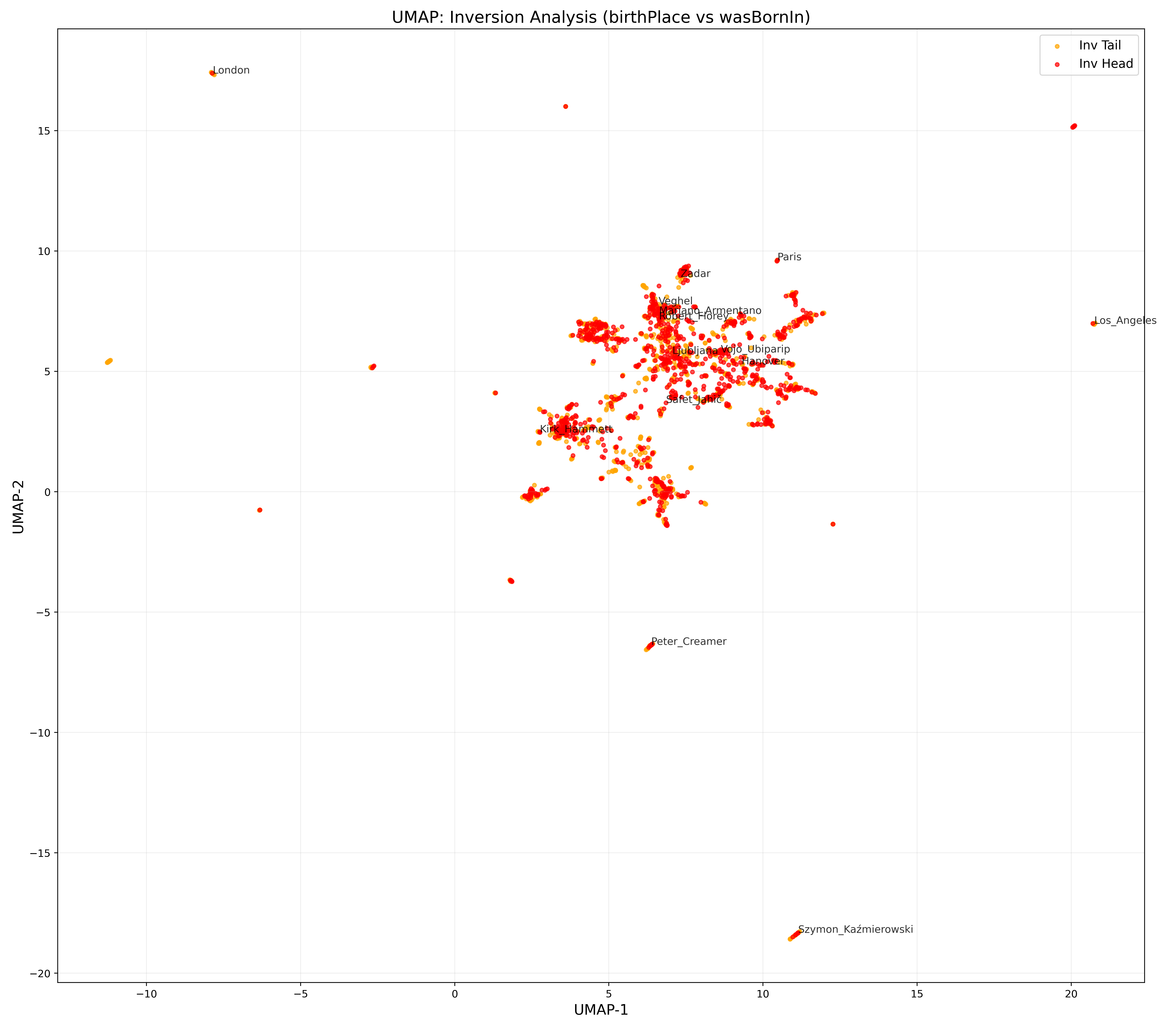}
  \caption{ \small \textbf{UMAP inversion analysis.} Embeddings affected by the \texttt{birthPlace} vs.\ \texttt{wasBornIn} perturbation show that \methodName preserves cluster integrity across head and tail roles, suggesting semantic continuity.}
  \label{fig:umap-inversion}
\end{figure}

Additionally, our design directly addresses long-standing limitations of previous models in modeling \emph{non-injective relations}. Both TransE and RotatE rely on injective geometric transformations (translations and rotations), which restrict their ability to represent many-to-one and one-to-many relation patterns commonly found in knowledge graphs~\cite{balazevic2019tucker, guo2020learning}. \methodName overcomes this by decoupling the embedding space into two semantically distinct subspaces: the modulus, which absorbs structural flexibility and topological variation, and the phase, which preserves directional semantics. The practical impact of this design is evident in Figure~\ref{fig:umap-inversion}, which visualizes entity embeddings under inversion perturbation (\texttt{birthPlace} $\leftrightarrow$ \texttt{wasBornIn}); \methodName preserves entity cluster cohesion across head and tail roles, maintaining semantic continuity even under adversarial inversion.

\begin{figure*}[t]
  \centering
  \begin{subfigure}[b]{0.48\textwidth}
    \centering
    \includegraphics[width=\linewidth]{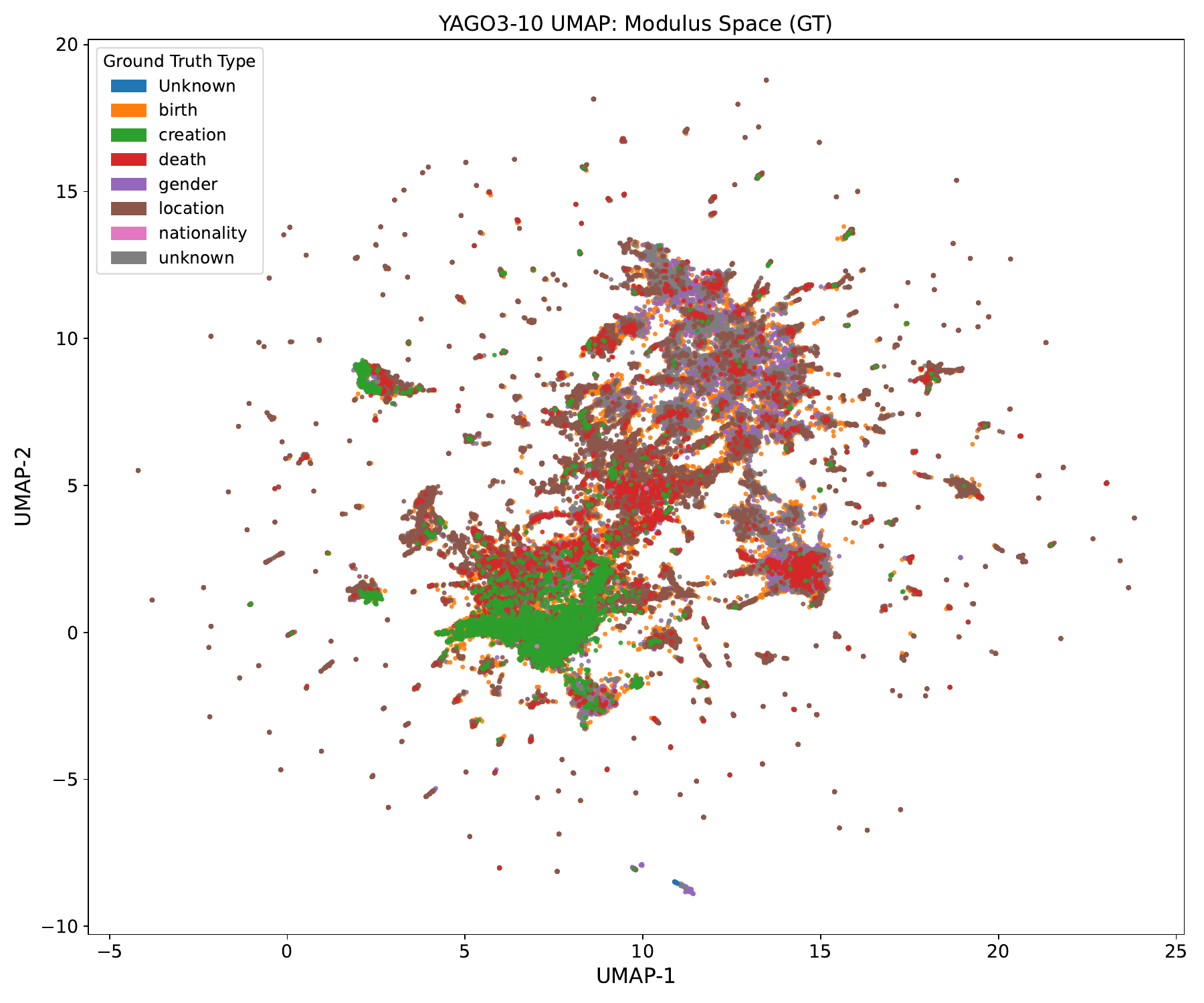}
    \caption{ \small UMAP of modulus embeddings. Clusters encode structural flexibility and reflect overlapping entity types.}
    \label{fig:umap-modulus}
  \end{subfigure}
  \hfill
  \begin{subfigure}[b]{0.48\textwidth}
    \centering
    \includegraphics[width=\linewidth]{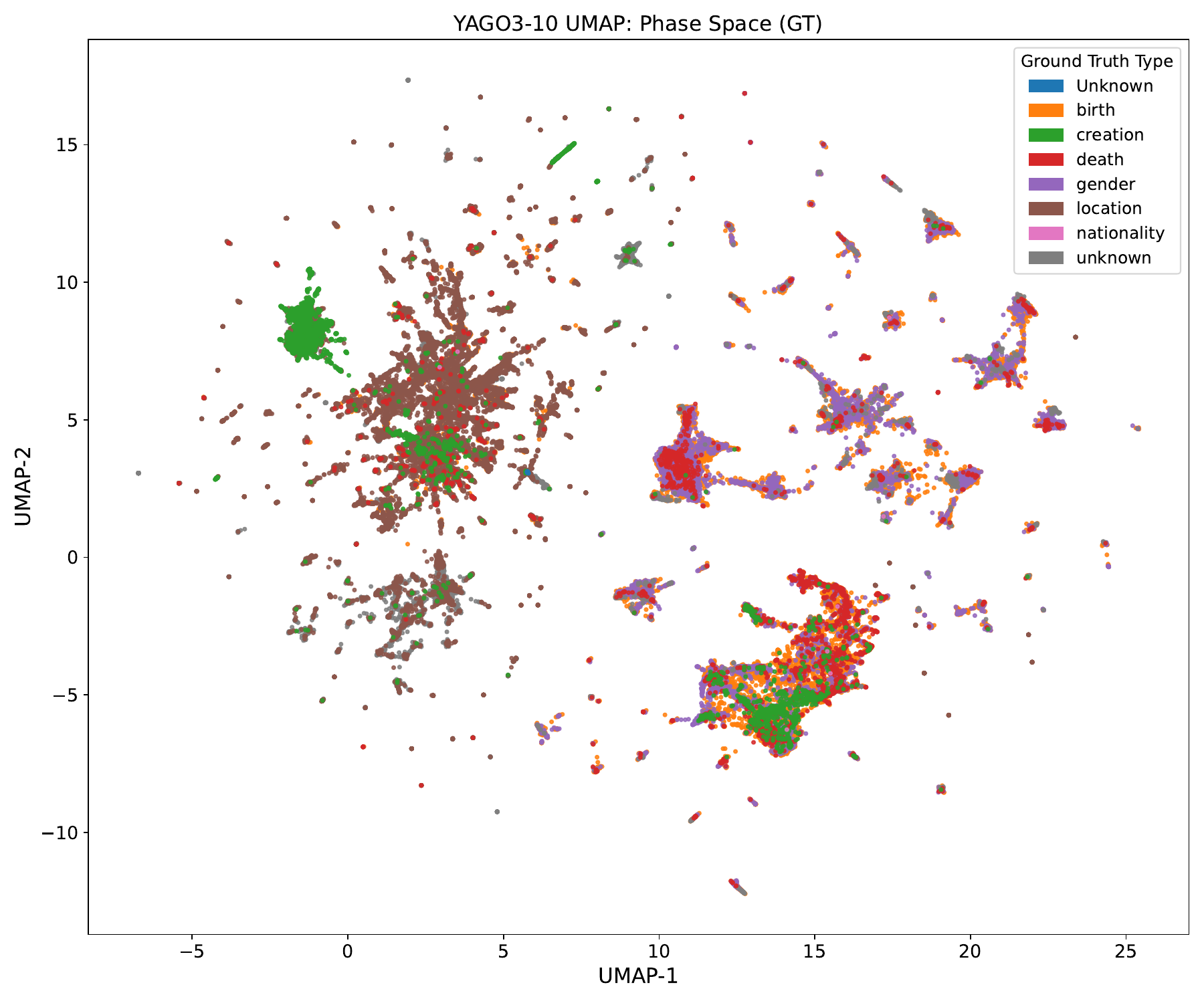}
    \caption{ \small UMAP of phase embeddings. Type-coherent clusters remain well-separated, preserving semantic structure.}
    \label{fig:umap-phase}
  \end{subfigure}
  \caption{ \small \textbf{RelatE’s phase and modulus decomposition on YAGO3-10.} The modulus space absorbs topological deformation, while phase embeddings retain directional semantic structure under perturbation.}
  \label{fig:umap-combined}
\end{figure*}

We further explore this architectural decomposition on YAGO3-10 (Figure~\ref{fig:umap-inversion}). The UMAP visualization of modulus embeddings (Figure~\ref{fig:umap-modulus}) shows flexible, diffuse clustering corresponding to dominant entity types ground truth, effectively absorbing structural noise. In contrast, phase embeddings (Figure~\ref{fig:umap-phase}) form tightly separated, type-coherent clusters, demonstrating that phase remains stable and interpretable even under perturbation. This modular separation is crucial: prior studies have identified the ``Z-paradox'' in TransE and RotatE~\cite{xu2020Zparadox}, where structurally distinct subgraphs become indistinguishable due to oversimplified embedding geometries. Our results suggest that \methodName mitigates this problem through explicit phase-modulus disentanglement.

In summary, \methodName not only matches or exceeds the benchmark performance of existing models but also delivers superior robustness and interpretability under conditions where previous architectures fail. These findings substantiate our core claim that real-valued embeddings, when properly structured, can ``go farther'' in knowledge base completion, bridging high task performance with architectural transparency and generalization under incomplete or noisy conditions.


\FloatBarrier
\section{Code Availability}
An anonymized GitHub repository with full training and evaluation scripts is available at:
\url{https://anonymous.4open.science/r/RelatE-36D5/}

\section{Broader Impact Statement}
\label{sec:broader_impact}
\methodName is a real-valued, interpretable model for knowledge graph completion. Its modular decomposition supports transparent reasoning and is particularly applicable to domains where interpretability and robustness are critical, such as healthcare, scientific discovery, and federated learning. By avoiding symbolic constraints, it broadens applicability to structured datasets underserved by prior models.

Risks include potential bias amplification from training data and misuse in automated decision systems. While \methodName offers geometric interpretability, it is not inherently robust to all data artifacts. Mitigations may include fairness-aware training or uncertainty modeling.

A promising future direction is integrating \methodName into large language model (LLM) workflows. Its real-valued structure aligns well with LLM latent representations, offering utility in retrieval-augmented generation, structured prompting, and neuro-symbolic reasoning toward more transparent, hybrid AI systems.

\end{document}